\documentclass{article}
\usepackage[dblblindworkshop,final]{neurips_2026}
\workshoptitle{VLM4RWD}

\usepackage[utf8]{inputenc}
\usepackage[T1]{fontenc}
\usepackage{graphicx}
\usepackage{booktabs}
\usepackage{amsmath}
\usepackage{amssymb}
\usepackage{microtype}
\usepackage{wrapfig}
\usepackage{multirow}
\usepackage{url}
\usepackage[hidelinks,hyperfootnotes=false]{hyperref}

\title{Selecting or Solving? What Best-of-$N$ Verification Delivers in Video Reasoning}

\author{%
  \bf Martin Q. Ma$^{1,2,}$\thanks{Work done while interning at Meta.} \quad
  Yuxiao Qu$^{2}$ \quad
  Ruslan Salakhutdinov$^{2}$ \\
  \bf Louis-Philippe Morency$^{2}$ \quad
  Paden Tomasello$^{1}$ \quad
  Juan Pino$^{1}$ \\[3pt]
  $^{1}$Meta \hspace{5em} $^{2}$Carnegie Mellon University
}

\begin{document}

\maketitle
\setcounter{footnote}{0}

\begin{abstract}
Answering a complex question about a long video is a grounding problem: the answer turns on a
few seconds of evidence somewhere inside it, and vision--language models (VLMs) remain well
below human accuracy at finding them. The standard way to spend extra test-time compute on this kind of video reasoning is best-of-$N$ selection: sample several chain-of-thought answers from a fixed model, then let a second model re-watch the video and pick one. Such pipelines are reported
by a single headline number, their margin over a majority vote. That number conflates two
effects with opposite deployment consequences: the benefit the second model adds by
\emph{selecting} among candidates, and the benefit it would have supplied by \emph{solving} the
question itself. We separate them, measuring the \emph{Selection Increment}
(SI)---selection accuracy minus the same verifier's accuracy with the candidates
withheld---alongside the margin over self-consistency, across five long-video benchmarks, with verifiers spanning weak open models to a frontier closed one. The two prove dissociable, for different reasons. The Selection Increment is
largest for the \emph{weakest} verifiers: a verifier that solves poorly on its own leaves the
candidates the most room to help. Beating the majority vote follows the opposite profile: where the vote already sits near its ceiling no verifier we test improves on it; below that ceiling only the strongest do. On Video-Holmes, the two come apart on a single
pool of candidates---a weak verifier gains $+19.3$ points from the candidates yet lands
$7.1$ points below the vote, while the frontier verifier gains the least of the five we
compare and beats the vote. The weakest verifier also shows a failure mode the capable
ones resist: on EgoSchema it picks by where an option sits
in the list rather than by what the video shows. Read as a merit score, SI puts a verifier that fails to beat a majority vote at the top of its ranking; read as an audit, it says which of the two effects a headline number delivers. Code is available at \url{https://github.com/facebookresearch/best-n-selection-video-reasoning}.
\end{abstract}

\begin{figure}[t]
    \centering
    \includegraphics[width=0.98\textwidth]{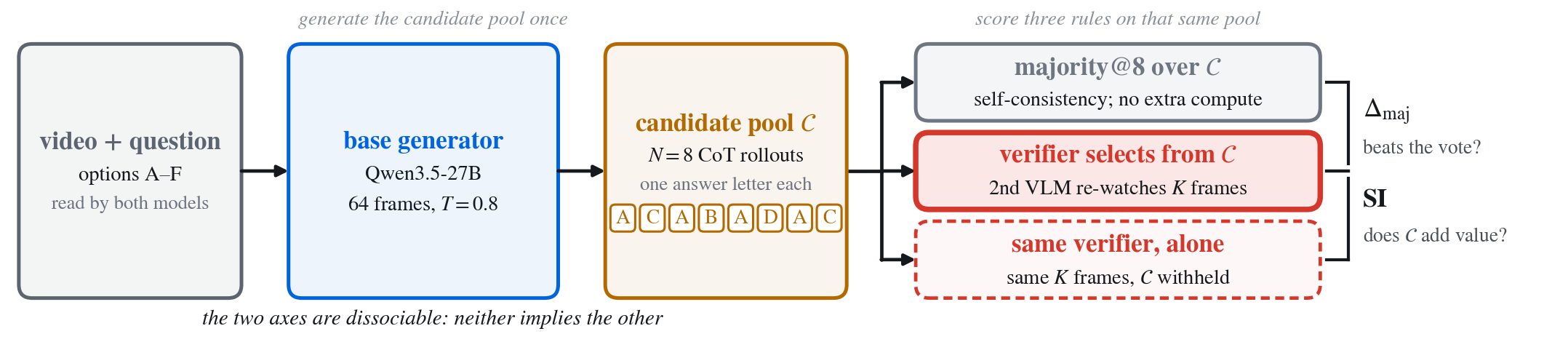}
    \caption{\textbf{What a best-of-$N$ verifier pipeline measures.} A base generator
    (Qwen3.5-27B) samples $N{=}8$ chain-of-thought rollouts per question, each
    ending in a candidate letter. Three rules score that same pool: majority voting; the verifier selecting among the candidates after re-watching $K$ frames; and the same verifier with the candidates withheld. Their differences define our two axes, $\Delta_{\text{maj}}$ and SI.}
    \label{fig:pipeline}
\end{figure}

\section{Introduction}
\label{section:intro}

A question about a long video is rarely answered by a summary. What decides it is a few
seconds of evidence buried somewhere in the footage, and a system is only right for the right
reason---only \emph{faithful} to the video---if its conclusion stays tied to that evidence.
Vision--language models (VLMs) now reason over video rather than describe it, yet on
benchmarks built to demand exactly this grounding, frontier models remain well below human
accuracy~\citep{cheng2025videoholmes,wang2024lvbench}. Video reasoning is therefore a
revealing setting for asking how additional inference-time compute should be spent: a
larger model is not the only lever. Our focus is the long-video form of the problem,
long-video question answering (long-video QA): complex questions, answered with
chain-of-thought, whose evidence is spread through long footage.

In video, that compute is most often spent \emph{gathering} evidence. Agentic loops
retrieve, caption, temporally ground, or zoom on
demand~\citep{zhang2023simple,wang2024videoagent,min2024morevqa,zhang2025vtimecot},
spending extra passes to surface the few relevant seconds. We study a less explored alternative that uses the same compute budget on \emph{deliberation}: sample several independent answers from a fixed
model and choose among them---best-of-$N$ selection, the recipe behind much of the gain
from test-time scaling in text
reasoning~\citep{cobbe2021training,lightman2023let,zheng2023judging}.

Sampling alone shows how much such selection could recover. On Video-Holmes, eight independent chains of thought from Qwen3.5-27B contain a correct
answer $78.8\%$ of the time, while self-consistency---majority voting over those chains---recovers only $58.9\%$. Close to twenty points are already present in the model's own output distribution and discarded by the vote, recoverable without retraining or a larger model. The natural way to claim them is a \emph{verifier}: a second VLM that re-watches the video and picks
one candidate (Figure~\ref{fig:pipeline}).

Gains reported for this recipe are read as evidence that the verifier selects well. They
cannot be. A verifier that re-watches the video is unavoidably also a second solver, so a single margin over the vote conflates the benefit added by \emph{selecting} among candidates
with the benefit the verifier would have supplied by \emph{solving} the question itself. Only the first adds something the base model could not produce; the second means the pipeline is a computationally expensive way to run a better model, and the right deployment differs in each case. Hence our question: \emph{when a second model appears to select well, is it selecting or merely solving---and which does a reported gain deliver?}

Answering it cleanly requires two controls, because the naive comparison flatters the verifier twice over. First, a model asked to pick can ignore the candidates and answer directly, and when that answer is correct, standard scoring still credits a successful selection; we therefore also score under a strict policy that only credits picks drawn from the candidates. Second, a strong verifier can beat self-consistency simply by answering well on its own. We therefore measure the \emph{Selection Increment} (SI): the verifier's accuracy when selecting among the candidates, minus its accuracy on the same questions with the candidates withheld. SI is the verifier's contribution \emph{net} of its own solving ability---exactly what a headline number leaves unresolved.

\paragraph{Findings and contributions.} With both controls in place, each pipeline is scored
on two questions---does the verifier add benefit beyond its own solving, and does it beat the
majority vote?---and our central finding is that the answers are \emph{dissociable}: neither
implies the other.
\begin{itemize}
\item \textbf{An audit protocol.} The two controls above---strict candidates-only scoring
and the Selection Increment---separate selecting from solving; both are computationally cheap, and neither
requires retraining (Section~\ref{section:setup}).
\item \textbf{Two axes, two levers.} Along a ladder of five verifiers over a fixed candidate
pool, SI falls monotonically as the verifier's own accuracy rises: the weakest verifier gains
the most from the candidates. Beating the vote is governed instead by two conditions, checked
in order---whether the vote already sits near its ceiling, and only then whether the verifier
is strong enough. An intuitive ``headroom'' rule---that selection helps most where the pool leaves the most room above the vote---finds no support (Section~\ref{section:rule}).
\item \textbf{Two failure modes of scaled selection.} Verifiers below a capability threshold
drift toward early answer positions rather than visual evidence, and letting the generator
gather evidence on demand does not raise the ceiling that selection works against
(Section~\ref{section:position}, Appendix~\ref{appendix:control}).
\item \textbf{What to report, and what to deploy.} A cross-model verifier does beat
self-consistency ($+7.94$ points on TempCompass); our strongest \mbox{Video-Holmes} margin, however, does not survive the strict scoring after multiplicity correction, so we do not claim it. The recommendation: report both axes, and the
scoring policy behind each (Section~\ref{section:bounds}).
\end{itemize}

\section{Related Work}
\label{section:related}

\paragraph{Best-of-$N$ verification and self-consistency.} Test-time compute can substitute
for parameters in text reasoning, through long chains of thought and sample-and-aggregate
schemes~\citep{brown2024large,muennighoff2025s1,wang2022self,deepseekai2025deepseek}.
Self-consistency, the majority vote over sampled chains~\citep{wang2022self}, is the
strongest training-free member of that family and our point of comparison throughout: any
verifier must beat it to justify the second model's computational cost. Selecting one candidate with a
verifier instead is equally well studied: outcome- and process-reward verifiers~\citep{cobbe2021training,lightman2023let}, the scaling of best-of-$N$ with sample
count~\citep{brown2024large}, LLM-as-a-judge evaluation~\citep{zheng2023judging}, and
generative verifiers that reason before scoring~\citep{zhang2024generative}. Two findings
from that literature shape our design: multi-model juries can outperform a single large
judge~\citep{verga2024replacing}, and judges prefer outputs from their own model
family~\citep{panickssery2024llm}. Nearer our setting, \citet{saluru2026disagreement} route
between Gemini and Claude on disagreement for implicit-reasoning video QA, and concurrent
work aligns multi-agent consensus to region-level visual evidence for image
VQA~\citep{wang2026seeing}. What this literature reports is the verifier's gain over
majority voting or pass@1, a quantity that conflates selection benefit with the verifier's own task skill. Our verifier is a judge applied listwise, making one holistic $N$-way pick;
we work on long video, ground selection in re-watched frames, and ask when selection adds
benefit at all.

\paragraph{Long video: gathering evidence, and auditing whether it helps.} In video the
dominant inference-time recipe gathers evidence with agentic loops that retrieve, caption,
temporally ground, or zoom, usually on small or medium
splits~\citep{wang2024videoagent,zhang2023simple,min2024morevqa,zhang2025vtimecot,janjua2026perception};
\citet{wang2026beacon} relate when such tool use helps to task difficulty. That line enriches
what the model sees; the pool of answers a verifier must choose among goes unmeasured, and
Appendix~\ref{appendix:control} finds it unmoved. A parallel line
makes the input frames query-conditioned rather than uniform, ranking the most relevant
frames per question at test
time~\citep{chen2026matryoshka,wang2026efficient,zou2026videobrain,he2026vtagent}; that axis is orthogonal to ours (it acts on the input, ours over sampled trajectories), so query-conditioned frames could feed both our generator and our verifier. We use uniform sampling throughout as a control. Running
alongside both lines is a growing audit literature asking whether reported gains reflect
added capability or format learning~\citep{guo2026do,mohamed2026cost} and showing that high
accuracy is often attainable without genuine visual grounding~\citep{lee2026accuracy};
closest to our task, \citet{alavi2026perception} find implicit video QA largely
perception-bound, with reasoning augmentation neutral to harmful. Our contribution belongs
to this audit tradition rather than the method-proposing one: we take a recipe that is
already standard and ask what its headline number is actually measuring.

\section{Measurement Protocol}
\label{section:setup}

\paragraph{Problem setup.} For a question $q$ with options $\mathcal{O}_q$, gold answer
$a^\star$, and video $V$, a base generator $\pi$ samples $N$ independent zero-shot
chain-of-thought trajectories $c_1,\dots,c_N\sim\pi(\cdot\mid V,q,\mathcal{O}_q)$, each
contributing a candidate letter $y_i=\mathrm{letter}(c_i)\in\mathcal{O}_q$; let
$\mathcal{C}=\{y_1,\dots,y_N\}$ be the candidate multiset. A selection rule maps
$\mathcal{C}$ to one answer. Our generator is Qwen3.5-27B~\citep{qwen2026qwen35} reading $64$ uniformly sampled frames ($32$ on EgoSchema) at temperature $T{=}0.8$, with $N{=}8$ for headline results and a sweep
$N\in\{1,2,4,8,16\}$ on Video-Holmes. The \emph{verifier} $J$ is a second VLM that
re-watches its own $K$ uniformly sampled frames $V_K$ and makes one holistic $N$-way pick,
$\hat y=J(V_K,q,\mathcal{O}_q,\mathcal{C})$ (Figure~\ref{fig:pipeline}). The same verifier can also be run without candidates, answering from its own frames alone,
$\hat y=J(V_K,q,\mathcal{O}_q)$; this is its \emph{standalone solver}, and it is what makes
the two axes separately measurable.

\paragraph{Two axes, and the headroom they live in.} We compare the verifier against
\emph{pass@1} (the expected accuracy of one chain drawn uniformly from $\mathcal{C}$),
\emph{self-consistency} $\hat y=\mathrm{mode}(\mathcal{C})$~\citep{wang2022self}, and the
best-of-$N$ \emph{oracle} (correct iff $a^\star\in\mathcal{C}$), which upper-bounds what any
selection rule can recover.\footnote{Majority-vote ties are broken by first appearance; the
rule is verified against the shipped implementation and is effectively letter-neutral
(Appendix~\ref{appendix:protocol}).} Writing $\mathrm{acc}(s)=\Pr[\hat y_s=a^\star]$,
\[
\Delta_{\text{maj}}=\mathrm{acc}(\text{sel})-\mathrm{acc}(\text{maj}),\qquad
\mathrm{SI}=\mathrm{acc}(\text{sel})-\mathrm{acc}(\text{standalone}),\quad
\mathrm{h}=\mathrm{acc}(\text{oracle})-\mathrm{acc}(\text{maj}).
\]
$\Delta_{\text{maj}}$ asks whether selection is worth its computational cost against the computationally cheap baseline;
the \emph{Selection Increment} $\mathrm{SI}$ asks whether the candidates help the verifier
beyond its own solving; and the headroom $\mathrm{h}$ bounds what any rule could recover.
The three are distinct, and the distinction is the paper's subject: a positive
$\mathrm{SI}$ means the candidates helped, but a weak standalone solver inflates it
(Section~\ref{section:rule}), so we report $\mathrm{SI}$ and $\Delta_{\text{maj}}$ together
throughout and never one alone.

\begin{table}[htbp]
    \centering
    \small
    \caption{\textbf{Base and verifier roles.} All \emph{headline} best-of-$N$ results fix
    Qwen3.5-27B as the base generator and vary the verifier; two base-varying
    $2\times2$s (Video-Holmes, TempCompass; Section~\ref{section:rule}) also vary the base.
    The standalone solver is the same verifier answering directly from its own $K$ frames.}
    \label{table:roles}
    \begin{tabular}{lll}
    \toprule
    Role & Model(s) & Frames \\
    \midrule
    Base generator ($N{=}8$ CoT, $T{=}0.8$)      & Qwen3.5-27B                   & 64 (32 on EgoSchema) \\
    Verifier, cross-model (headline)             & Muse Spark 1.1                & 48 \\
    Verifier, cross-model (strong)               & GPT-5.6 Sol                   & 32 \\
    Verifier, same-family                        & Qwen3.5-27B / -397B           & 48 / 48, 96 \\
    Verifier, weak cross-family                  & Phi-3.5-V                     & 16, 24, 48, 64 \\
    Verifier, mid cross-family                   & InternVL3-8B / -38B           & 48 / 24, 48 \\
    \bottomrule
    \end{tabular}
\end{table}

\paragraph{Verifiers and benchmarks.} To trace how selection depends on verifier strength we
vary the verifier along model family and capability (Table~\ref{table:roles}), from a weak
cross-family model, Phi-3.5-V~\citep{abdin2024phi3}, through two mid-scale open ones,
InternVL3-8B and -38B~\citep{zhu2025internvl3}, and a same-family sibling $15\times$ the
base's size, Qwen3.5-397B~\citep{qwen2026qwen35moe}, up to two frontier \emph{cross-model} verifiers, a reasoning VLM from another family (Muse Spark 1.1~\citep{metaai2026musespark}, hereafter Muse) and the stronger closed GPT-5.6 Sol, a later release in the GPT-5 family~\citep{openai2025gpt5card}, over compute budgets
$K\in\{16,24,32,48,64,96\}$. Every headline
result fixes Qwen3.5-27B as generator and varies the verifier; two base/verifier swaps that
also vary the base are reported in Section~\ref{section:rule}. We measure on five video
multiple-choice QA benchmarks: our primary reasoning anchor
Video-Holmes~\citep{cheng2025videoholmes} ($1{,}837$ six-way reasoning MCQs over $270$
videos), with TempCompass~\citep{liu2024tempcompass}, MLVU-Test~\citep{zhou2024mlvu},
VRBench~\citep{yu2025vrbench}, and EgoSchema~\citep{mangalam2023egoschema}. They are chosen so self-consistency accuracy spans a wide range, from about $59\%$ (Video-Holmes) to about $91\%$ (VRBench), so selection is tested in easy and hard regimes alike. All deltas are
matched-sample, computed on the intersection of items with valid predictions from both
compared methods; split sizes and per-cell denominators are in
Appendix~\ref{appendix:protocol}.

\paragraph{Scoring: selecting versus solving.} A verifier asked to select can solve instead: it may emit a letter outside the candidate set, and when that out-of-candidate
(OOC) letter is correct, naive scoring credits a successful selection. We therefore score
every cell under three policies: \emph{Solver} counts OOC hits (the default in prior work),
\emph{Selector} counts a pick only if it is correct \emph{and} drawn from $\mathcal{C}$, and
\emph{Snap} maps an OOC pick to its nearest candidate. Headline numbers use \emph{Solver} so
they remain comparable to published best-of-$N$ results; we report \emph{Selector} as a
labeled ablation and flag every disagreement, since the gap between the two is exactly the
leakage a headline number over-credits. The standalone leg of $\mathrm{SI}$ is
policy-invariant, because the solver never sees $\mathcal{C}$. Per-cell OOC rates, which run
from under a percent to roughly $13\%$ for a weak verifier, are in
Appendix~\ref{appendix:protocol}.

\paragraph{Statistics.} Every headline delta is matched-sample and tested with an exact
two-sided paired McNemar test on the discordant pairs. We report per-comparison significance
and flag non-significant deltas (n.s.) throughout, applying Holm--Bonferroni correction to
the flagship beats-majority family where it is named and treating isolated significant
results in the broader uncorrected grid accordingly. Permutation and Fisher tests in
Section~\ref{section:position} are one-sided against a predicted direction, a deliberate
seam with the two-sided McNemar tests that we state rather than smooth over
(Appendix~\ref{appendix:protocol}).

\section{Two Axes, Two Levers}
\label{section:beats}
\label{section:rule}

\begin{wraptable}{r}{0.36\textwidth}
    \vspace{-1.1\intextsep}
    \centering
    \small
    \caption{\textbf{Video-Holmes} ($1{,}837$ six-way MCQs), $N{=}8$. Best verifier: GPT-5.6 Sol.}
    \label{table:vhholmes}
    \begin{tabular}{lr}
    \toprule
    Rule & acc.\ (\%) \\
    \midrule
    pass@1           & $55.6$ \\
    self-consistency & $58.9$ \\
    best verifier    & $62.0$ \\
    oracle@8         & $78.8$ \\
    \bottomrule
    \end{tabular}
    \vspace{-\intextsep}
\end{wraptable}
\paragraph{On our primary anchor, only a strong verifier beats self-consistency.}
Video-Holmes gives the cleanest reading, because there is a great deal to win: sampling
$N{=}8$ chains from the $27$B base yields a self-consistency accuracy of $58.9\%$ against a
best-of-$N$ oracle of $78.8\%$, a $\approx20$-point selectable gap that widens with $N$
(at $N{=}16$ the oracle reaches $82.5\%$ while the vote stays flat at $59.0\%$;
Table~\ref{table:vhholmes}). The strongest verifier we run, GPT-5.6 Sol, selects at
$62.0\%$---$+6.4$ over pass@1 and $+3.1$ over self-consistency (exact McNemar
$p{=}7.0\times10^{-4}$)---while still leaving $\approx17$ points of oracle headroom on the
table. The four weaker verifiers all fall \emph{below} self-consistency. Selection helps
here, but only with a verifier strong enough to be worth its own computational / inference cost: the
paper's thesis in miniature.\footnote{Per-cell denominators, and the check that scoring unanswered items wrong rather than excluding them leaves every conclusion unchanged, are in Appendix~\ref{appendix:protocol}.}

\begin{table}[t]
    \centering
    \small
    \caption{\textbf{The two axes come apart on full splits.} Solver-policy (prior-work convention)
    accuracies for the Muse holistic-48 and GPT-5.6 Sol verifiers; majority is the
    shared $27$B majority@8 baseline. The headroom $\mathrm{h}$ is a per-benchmark property of the candidate pool and predicts neither delta. VRBench cells use the first 1{,}000 questions of its split ($118$ of $960$ videos); deltas are McNemar-significant at $\alpha{=}0.05$ unless marked n.s.}
    \label{table:headline}
    \begin{tabular}{lrrrrrr}
    \toprule
    Benchmark & $n$ & Majority (\%) & Selection (\%) & $\Delta_{\text{maj}}$ (pt) & SI (pt) & h (pt) \\
    \midrule
    TempCompass \emph{(Muse)}        & 1{,}574 & 79.2 & 87.2 & $+7.94$ & $-1.08$ & \multirow{2}{*}{$\approx11$} \\
    TempCompass \emph{(GPT-5.6 Sol)} & 1{,}570 & 79.1 & 88.5 & $+9.43$ & $+1.15$ & \\
    \addlinespace[2.5pt]
    MLVU-Test \emph{(Muse)}        & 496 & 73.2 & 74.0 & $+0.81$\,\textsuperscript{n.s.} & $+5.68$ & \multirow{2}{*}{$\approx14$} \\
    MLVU-Test \emph{(GPT-5.6 Sol)} & 502 & 72.9 & 70.9 & $-1.99$\,\textsuperscript{n.s.} & $+11.16$ & \\
    \addlinespace[2.5pt]
    VRBench \emph{(Muse)}        & 983 & 91.0 & 87.8 & $-3.26$ & $+3.67$ & \multirow{2}{*}{$\approx6$} \\
    VRBench \emph{(GPT-5.6 Sol)} & 989 & 90.9 & 89.3 & $-1.62$ & $+3.96$ & \\
    \bottomrule
    \end{tabular}
\end{table}

\paragraph{Across the suite, the two axes come apart.} Table~\ref{table:headline} reports
both axes for the two deployed cross-model verifiers on full splits. The margin over
self-consistency shrinks as the candidate pool saturates: on TempCompass ($n{=}1{,}574$)
the cross-model verifier reaches $87.2\%$ against $79.2\%$, a gain of $7.94$ points and its
largest anywhere; on MLVU-Test it only ties ($+0.81$, n.s.); and on VRBench, whose vote already reaches $91\%$, it loses ($-3.26$, $p{=}5.6\times10^{-6}$). The Selection Increment moves independently of that margin, and sometimes against it: SI is positive on
MLVU-Test ($+5.68$) and VRBench ($+3.67$) precisely where the verifier fails to beat the
vote, and \emph{negative} on TempCompass ($-1.08$) where the verifier wins by almost eight points. There, Muse scores higher answering directly than selecting. A reader given only the margin would conclude selection works on TempCompass and not on VRBench; given only SI, the opposite---which is why we report the pair.

\paragraph{The intuitive ``headroom'' rule finds no support.} The natural predictor of
selection benefit is the room the sampler leaves, $\mathrm{h}=\text{oracle}-\text{majority}$.
Individual benchmarks violate any monotone version of the rule in \emph{both} directions: on TempCompass, at low headroom ($\approx11$ points), the verifier beats the vote by $+9.4$ ($p{<}10^{-23}$), whereas on MLVU-Test, at higher headroom ($\approx14$ points), it underperforms the vote ($-2.0$, n.s.). Across a compressed $\approx6$--$14$-point range we claim no reliable
correlation in either direction, only that there is no evidence for the rule. Headroom is not the \emph{level} of the majority baseline: one is a gap, the other a position. Only the second carries signal below.

\paragraph{Lever one: the increment tracks standalone accuracy, inversely.} Holding the
candidate pool fixed and sweeping verifiers of increasing standalone accuracy drives SI
monotonically toward zero. A five-verifier ladder on the Video-Holmes Qwen-base pool makes
this explicit: SI falls from $+28.7$ (Qwen-27B self-verifier, standalone $27.7\%$) through
$+19.3$ (Phi-3.5, $32.5\%$), $+17.6$ (InternVL3-8B, $39.7\%$) and $+10.9$ (InternVL3-38B,
$47.7\%$) to $+7.9$ (GPT-5.6 Sol, $54.1\%$), and an MLVU-Test ladder reproduces the
ordering ($+26.3$, $+21.6$, $+20.8$, $+10.8$ as standalone rises
$38.6\!\to\!46.7\!\to\!51.1\!\to\!62.0\%$).\footnote{The two ladders share a rung and are not independent, and one rung runs at a smaller frame budget than its siblings; Appendix~\ref{appendix:provenance} states all three provenance notes in full.} A large
increment therefore marks a weak \emph{solver} that can nonetheless select. SI measures how much the candidates add to a given verifier, not how good that verifier is. The one swap
that appears to run the other way confirms the rule: replacing Muse with GPT-5.6 Sol raises
SI on every deployed benchmark, not because GPT selects better but because it
\emph{under-solves} those benchmarks by more, leaving the candidates more room
(Appendix~\ref{appendix:extended}).

\paragraph{The two axes dissociate inside a single candidate pool.}
\begin{wrapfigure}{r}{0.52\textwidth}
    \vspace{-1.0\intextsep}
    \centering
    \includegraphics[width=0.52\textwidth]{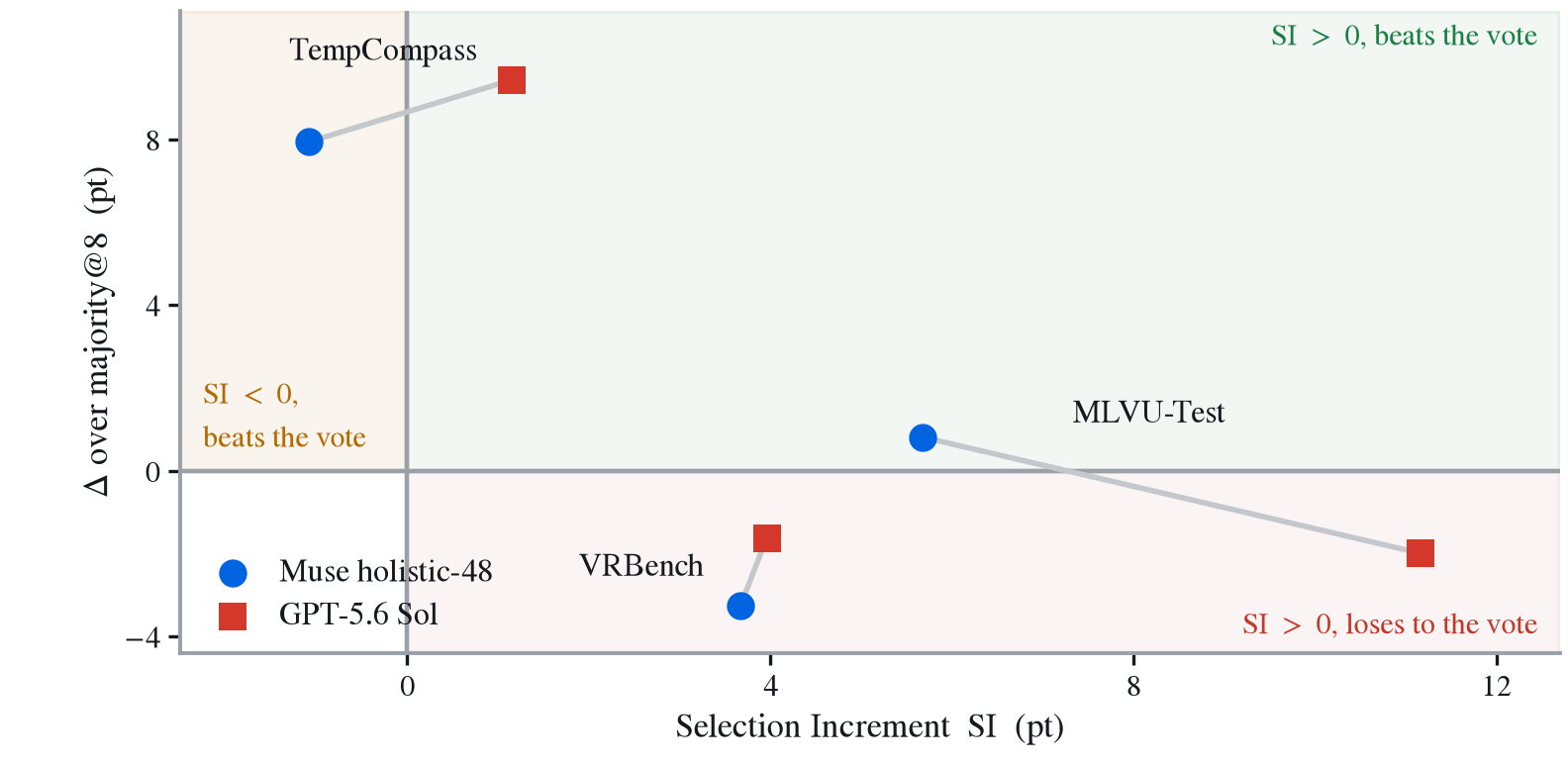}
    \caption{\textbf{Selection Increment vs.\ gain over
    self-consistency.} Each point is a (benchmark,
    verifier) pair; lines link the two verifiers on one
    benchmark. SI$>$0 and beating self-consistency are
    dissociable. Solver policy; VRBench: first 1{,}000
    questions ($118$ of $960$ videos).}
    \label{fig:taxonomy}
    \vspace{-1.1\intextsep}
\end{wrapfigure} On the Video-Holmes
Qwen-base pool, base difficulty is held fixed and the verifier alone varies, which yields
the paper's sharpest contrast (Table~\ref{table:vh2x2}). The weak Phi-3.5 verifier selects
far better than it solves---SI $+19.3$ from a $32.5\%$ standalone---yet its selected
accuracy lands $7.1$ points \emph{below} the majority vote ($51.8\%$ vs.\ $58.9\%$; exact
McNemar $p{<}10^{-13}$), while the strong Muse verifier adds almost nothing over its own
solving ($+2.5$) and significantly beats the vote ($+2.8$;
$p{=}2.3\times10^{-3}$). Neither axis predicts the other, on one and the same pool. Three
controls place the weak verifier's deficit in its capability rather than in this pool or its
protocol: it reproduces on MLVU-Test against a different solver pool and a different
baseline ($8.0$ points below the vote, exact McNemar $p{=}2.4\times10^{-6}$); re-running the
verifier at the generator's own $64$ frames leaves both deficits numerically larger rather than closing them, which falsifies the frame-handicap account without establishing
what the real mechanism is; and swapping only verifier strength at a common $24$-frame budget moves the outcome from that $8.0$-point deficit to statistical parity with
the vote (InternVL3-38B, $\Delta_{\text{maj}}{=}-0.2$, exact McNemar $p{=}1.000$). The
pathology is therefore not confounded with frame budget, model family, or parameter count
(Appendix~\ref{appendix:extended}).\footnote{The verifier's frame budget is not a lever in
either direction: across six paired comparisons no rung improves selection, and the one rung
that isolates frames under an otherwise identical engine configuration is null on both
benchmarks (Appendix~\ref{appendix:extended}).}

\begin{table}[t]
    \centering
    \small
    \caption{\textbf{Base $\times$ verifier on Video-Holmes (full $2\times2$).} Selection Increment (Panel~A) and gain over self-consistency (Panel~B, selection vs.\ majority@8), in points, for a weak and a strong verifier (Qwen-27B, standalone $27.7$; Muse, $59.4$) over a Qwen- and a Muse-generated candidate pool (each split $n{=}1{,}837$;
    the Muse-verifier cells use matched $n{=}1{,}808$/$1{,}807$). SI is a \emph{column} effect; beating self-consistency is set by the \emph{verifier}.
    Cell-level tests and provenance: Appendix~\ref{appendix:extended}.}
    \label{table:vh2x2}
    \begin{tabular}{lcc}
    \toprule
     & Qwen-27B verifier & Muse verifier \\
    Base pool & \emph{(weak, std.\ $27.7$)} & \emph{(strong, std.\ $59.4$)} \\
    \midrule
    \multicolumn{3}{l}{\emph{Panel A: Selection Increment (SI $=$ selection $-$ standalone)}} \\
    Qwen-base (split $n{=}1{,}837$) & $+28.7$ & $+2.5$ \\
    Muse-base (split $n{=}1{,}837$) & $+31.2$ & $+0.9$ \\
    \midrule
    \multicolumn{3}{l}{\emph{Panel B: gain over self-consistency ($\Delta_{\text{maj}} =$ selection $-$ maj@8)}} \\
    Qwen-base & $+0.05$ ($58.95$ vs.\ $58.90$) & $+2.8$ ($61.9$ vs.\ $59.1$) \\
    Muse-base & $+0.00$ ($59.55$ vs.\ $59.55$) & $+0.8$ ($60.3$ vs.\ $59.5$) \\
    \bottomrule
    \end{tabular}
\end{table}

\paragraph{Lever two: the saturation gate first, then the strength gate.}
Ordering every base$\times$verifier cell
by the majority accuracy of its pool separates two conditions. Above roughly $87\%$ the \emph{saturation gate} decides and strength is irrelevant: on the Muse-base TempCompass pool ($87.5\%$)
and on VRBench ($\approx91\%$) no verifier beats the vote, and on VRBench both deployed verifiers significantly lose: frontier GPT-5.6 Sol ($-1.62$) no less than Muse ($-3.26$).
Below the ceiling the \emph{strength gate} decides: the two strongest verifiers win on the Qwen-base
Video-Holmes pool ($58.9\%$; $+2.8$ and $+3.1$) and every weaker one fails to. On the Qwen-base TempCompass pool ($79.2\%$) the weak self-verifier only draws level ($+0.9$, exact McNemar $p{=}0.082$)---a null with a mechanism: it agrees with the plurality on $95.1\%$
of rows, rubber-stamping the modal candidate rather than exercising independent judgment.
Because the TempCompass $2\times2$ holds the task fixed and varies the pool, it isolates
saturation the way Video-Holmes isolates strength: the Muse verifier's $+7.94$ win
collapses to a null ($-0.4$) as majority rises from $79.2$ to $87.5$
(Table~\ref{table:swap2x2}). The two gates are necessary rather than jointly sufficient: MLVU-Test sits in the permissive range ($\approx73\%$) with a capable verifier, and still neither deployed verifier beats the vote ($+0.81$ and $-1.99$, both n.s.)---a null we cannot presently explain and do not paper over. And the ceiling itself rests on two near-saturated pools, so we report an observed boundary rather than a calibrated threshold.
Figure~\ref{fig:taxonomy} places every deployed pair in these two coordinates.

\paragraph{Gathering more evidence does not raise the ceiling.} One reading of these limits is
that selection looks weak only because the sampled candidates are impoverished, and that
letting the model gather more evidence would lift the ceiling a verifier works against.
It does not. Augmenting \emph{candidate generation} with an on-demand agentic loop that may
zoom, crop, retrieve frames, or ground objects leaves the best-of-$N$ oracle on EgoSchema unchanged ($\Delta_{\text{oracle}}{=}0.000$ at $n{=}500$), and giving the same loop
to a strong cross-model verifier leaves its accuracy within noise of its zero-shot self.
Selection can only surface a correct answer the sampler already produced, which is what
directs this study at selection rather than gathering (Appendix~\ref{appendix:control}).

\section{A Failure Mode of Under-Capable Verifiers: Answer-Position Bias}
\label{section:position}

Answer-position bias has as a threshold:
below a capability floor a verifier drifts toward earlier options and away from the last; above that floor the early drift is absent. A verifier choosing among lettered options can appear to discriminate while favoring an early slot whenever its grounding is too weak to tell the candidates apart, in which case part of what we credit to selection is an artifact of letter position rather than of visual evidence. We quantify this two ways. \emph{A-lift} is
the rate at which a verifier picks ``A'' minus the rate at which ``A'' is gold, in points;
\emph{primacy} is the mean normalized position of the gold answer minus that of the pick, so
a positive value means the verifier systematically chooses earlier options than gold
warrants. Primacy captures the full early-to-late slope rather than one letter, and the two
agree closely across our $22$ verifier--benchmark cells (Spearman $+0.942$).

Bias falls with standalone accuracy, but discontinuously. The rank correlation between
accuracy and A-lift is $\rho{=}-0.954$ across the ten configurations (exact permutation
$p{=}4.5\times10^{-5}$ over all $10!$ label assignments), and sweeping the $27$B
self-verifier's two ranks over the whole range consistent with a corrected, higher standalone
accuracy leaves $\rho$ at $-0.73$ or stronger. The under-capable end carries that ordering
entirely: dropping the weak Phi-3.5-Vision verifier moves the pooled within-benchmark test
from $p{=}0.013$ to $p{=}0.34$, and among capable verifiers alone there is no accuracy-to-bias
trend. Of the $22$ cells, nine are individually significant by an exact McNemar test and five
survive Holm--Bonferroni correction; all four \emph{positive} survivors belong to
Phi-3.5-Vision, while the fifth is the capable Qwen3.5-27B self-verifier sitting \emph{below}
gold on Video-Holmes ($-2.8$ points, $p{=}0.0020$, Solver policy)---a signed control rather
than a bias of its own. The honest description is a threshold.

EgoSchema is the sharpest test, because it carries a biased arm and a control arm on
the same questions. It is five-way with near-uniform gold whose \emph{mode is E} (gold-``A''
$20.2\%$, gold-``E'' $23.4\%$; $n{=}500$), so an early-option default \emph{lowers}
accuracy: any bias it reveals is a liability, not free accuracy. Only the genuinely weak Phi-3.5-Vision shows a clear default: under the Solver policy, A-lift $+4.4$ at $16$ frames and $+8.0$ at $48$ (Table~\ref{table:position}), with the modal answer E correspondingly under-selected ($13.4\%$ picked vs.\ $23.4\%$ gold). Every capable verifier stays within $1.4$ points of gold. The failure mode is a graded positional slope rather than a
first-option collapse, and recency avoidance---the deficit at the \emph{last}
option---persists at every capability level, including the frontier, where the first-option excess is absent (Appendix~\ref{appendix:extended}).

\textbf{The significant bias is confined to a degenerate response mode.} Instrumenting the weak
verifier's raw output localizes the preference. Its responses fall into two disjoint modes,
separated perfectly by output length: a \emph{marker} mode returning the requested
\texttt{Answer:\,X} form, and a \emph{terse} mode returning a bare letter and nothing
else. Neither truncates: every response terminates normally, and a bare letter is a complete, unambiguous verdict. What follows is therefore the verifier's behavior rather than an artifact of how we parse it. The significant A-lift lives entirely in the terse mode: across ten
verifier--benchmark arms, marker-mode responses show no significant A-lift anywhere ($10$ of
$10$), while terse-mode responses favor the first option in every arm where that mode occurs
($8$ of $8$, $+6.7$ to $+32.4$ points, largest $p{=}2.5\times10^{-3}$; all eight survive
Holm--Bonferroni over the full family of eighteen mode tests). Nor is it accuracy in
disguise. On EgoSchema at $16$ frames the two modes are equally accurate to within $0.11$
points and the terse mode is still biased; on MLVU-Test and Video-Holmes the terse mode is the
\emph{less} accurate one, by $6.8$ to $16.7$ points, which cannot manufacture a preference for
the first option either. The default is directional rather than a by-product of solving
worse. Because that partition sits \emph{within} one verifier at a fixed
model and frame budget, what varies across it is the mode and not the model. A capable
verifier that never enters the terse mode adds a further control---InternVL3-38B emits the
marker on all $502$ items of both arms and reaches significance in neither---but mode and
strength are collinear there, so it rules out extraction failure and item difficulty rather
than verifier strength (Appendix~\ref{appendix:extended}). The mode structure revises the scope of the mechanism,
not the deficit it explains: all twelve $\Delta_{\text{maj}}$ cells are negative and
significant under both scored fields, so a weak verifier falls below the vote
whether or not it is in the mode that biases it.

\section{Verifying versus Solving: What to Report, What to Deploy}
\label{section:bounds}

Where the Selection Increment is not positive, selection provides only accuracy the
verifier already gets by solving---at a higher computational cost. On TempCompass with the Muse verifier
($\mathrm{SI}{=}-1.08$) the same verifier used for standalone solving is more accurate than
the pipeline, and it avoids the computational cost of reading the $27$B candidates, so verification there is
the wrong way to spend the verifier's compute. Where $\mathrm{SI}>0$ (MLVU-Test with Muse, or TempCompass with the stronger GPT-5.6 Sol), selection does beat standalone solving and the pipeline earns its extra computational cost. A benchmark-matched token accounting is left to future work, so
we make no blanket compute-efficiency claim for verification; Appendix~\ref{appendix:cost}
sets out the reasoning in passes rather than tokens.

The sign of the increment is also what a reader needs in order to interpret someone else's
headline. A reported gain over majority voting is silent on which of the two systems of
Section~\ref{section:rule} produced it, and our suite contains both: a weak verifier that
selects far better than it solves and still falls below the vote, and a strong verifier
that beats the vote while adding almost nothing to its own solving. Read as a merit score, SI's top-ranked verifier is one that fails to beat a vote; read as an audit,
it says which of the two effects a reported gain delivers, and that is how we intend it. The
reading is deflationary---on our anchor, the verifier worth deploying is the one that adds
the \emph{least} selection benefit.

Both axes are computationally cheap enough to check before committing inference compute. On a labeled development slice, majority accuracy over the sampled candidates does not require additional compute once the candidates exist, so the saturation gate can be evaluated before any verifier is called; the increment requires one extra pass, the verifier answering with the candidates withheld. That makes the failure predictable in advance,
and yields a three-step allocation policy.
First, skip verification where the vote already sits near the observed ceiling: no verifier in our study beat it there. Second, below that ceiling, pair the base with a verifier clearly
stronger than itself and drawn from a different model family, so that its errors are
decorrelated from the base's. Third, where even that verifier does not exceed the vote, do not allocate additional compute for selection: deploy whichever of the two computationally cheaper alternatives the protocol has already measured scores higher, the majority vote or the verifier answering alone. On TempCompass with Muse the standalone verifier wins that comparison; on MLVU-Test the vote does.

\section{Conclusion}
\label{section:conclusion}
When a verifier beats the majority vote, the margin alone does not say why: the verifier may be selecting well, or simply solving the question better than the base model. Two cheap controls—strict candidates-only scoring and the Selection Increment—tell these apart, and they show that the two things a pipeline can do are separable. Across five verifiers on a fixed candidate pool, the weakest verifier gained the most from seeing candidates, while beating the vote depended on how close the vote already was to its ceiling and, only then, on verifier strength. Weak verifiers also latched onto answer position instead of the video, and letting the generator gather more evidence did not raise the ceiling. Reported on both axes, genuine selection gains do appear—a cross-model verifier adds $+7.94$ points over self-consistency on TempCompass—and we see this stricter accounting as the way to find more of them.

\begin{ack}
This project is funded by Meta AI mentorship program. We appreciate the earlier discussions with Rongjie Huang and Jason Zheng, the great help Hongyu Gong provided for open-sourcing the code, and the help from Harris Warren and other members of C\&L team in FAIR for getting this project approved.
\end{ack}

\bibliographystyle{assets/plainnat}
\bibliography{paper}

@article{brown2024large,
  title  = {Large Language Monkeys: Scaling Inference Compute with Repeated Sampling},
  author = {Bradley Brown and Jordan Juravsky and Ryan Ehrlich and Ronald Clark and Quoc V. Le and Christopher Re and Azalia Mirhoseini},
  journal= {arXiv preprint arXiv:2407.21787},
  year   = {2024}
}

@article{deepseekai2025deepseek,
  title  = {DeepSeek-R1: Incentivizing Reasoning Capability in LLMs via Reinforcement Learning},
  author = {DeepSeek-AI},
  journal= {Nature},
  volume = {645},
  pages  = {633--638},
  note   = {arXiv:2501.12948},
  year   = {2025}
}

@inproceedings{wang2024videoagent,
  title  = {VideoAgent: Long-form Video Understanding with Large Language Model as Agent},
  author = {Xiaohan Wang and Yuhui Zhang and Orr Zohar and Serena Yeung},
  booktitle = {European Conference on Computer Vision},
  note   = {arXiv:2403.10517},
  year   = {2024}
}

@inproceedings{min2024morevqa,
  title  = {MoReVQA: Exploring Modular Reasoning Models for Video Question Answering},
  author = {Juhong Min and Shyamal Buch and Arsha Nagrani and Minsu Cho and Cordelia Schmid},
  booktitle = {Proceedings of the IEEE/CVF Conference on Computer Vision and Pattern Recognition},
  note   = {arXiv:2404.06511},
  year   = {2024}
}

@inproceedings{zhang2023simple,
  title  = {A Simple LLM Framework for Long-Range Video Question-Answering},
  author = {Ce Zhang and Taixi Lu and Md Mohaiminul Islam and Ziyang Wang and Shoubin Yu and Mohit Bansal and Gedas Bertasius},
  booktitle = {Proceedings of the 2024 Conference on Empirical Methods in Natural Language Processing},
  note   = {arXiv:2312.17235},
  year   = {2024}
}

@inproceedings{muennighoff2025s1,
  title  = {s1: Simple test-time scaling},
  author = {Niklas Muennighoff and Zitong Yang and Weijia Shi and Xiang Lisa Li and Li Fei-Fei and Hannaneh Hajishirzi and Luke Zettlemoyer and Percy Liang and Emmanuel Candes and Tatsunori Hashimoto},
  booktitle = {Proceedings of the 2025 Conference on Empirical Methods in Natural Language Processing},
  note   = {arXiv:2501.19393},
  year   = {2025}
}

@inproceedings{wang2022self,
  title  = {Self-Consistency Improves Chain of Thought Reasoning in Language Models},
  author = {Xuezhi Wang and Jason Wei and Dale Schuurmans and Quoc Le and Ed Chi and Sharan Narang and Aakanksha Chowdhery and Denny Zhou},
  booktitle = {International Conference on Learning Representations},
  note   = {arXiv:2203.11171},
  year   = {2023}
}

@inproceedings{mangalam2023egoschema,
  title  = {EgoSchema: A Diagnostic Benchmark for Very Long-form Video Language Understanding},
  author = {Karttikeya Mangalam and Raiymbek Akshulakov and Jitendra Malik},
  booktitle = {Advances in Neural Information Processing Systems 36},
  note   = {arXiv:2308.09126},
  year   = {2023}
}

@inproceedings{lightman2023let,
  title  = {Let's Verify Step by Step},
  author = {Hunter Lightman and Vineet Kosaraju and Yura Burda and Harri Edwards and Bowen Baker and Teddy Lee and Jan Leike and John Schulman and Ilya Sutskever and Karl Cobbe},
  booktitle = {International Conference on Learning Representations},
  note   = {arXiv:2305.20050},
  year   = {2024}
}

@inproceedings{liu2024tempcompass,
  title  = {TempCompass: Do Video LLMs Really Understand Videos?},
  author = {Yuanxin Liu and Shicheng Li and Yi Liu and Yuxiang Wang and Shuhuai Ren and Lei Li and Sishuo Chen and Xu Sun and Lu Hou},
  booktitle = {Findings of the Association for Computational Linguistics ACL 2024},
  note   = {arXiv:2403.00476},
  year   = {2024}
}

@inproceedings{zhou2024mlvu,
  title  = {{MLVU}: Benchmarking Multi-task Long Video Understanding},
  author = {Junjie Zhou and Yan Shu and Bo Zhao and Boya Wu and Zhengyang Liang and Shitao Xiao and Minghao Qin and Xi Yang and Yongping Xiong and Bo Zhang and Tiejun Huang and Zheng Liu},
  booktitle = {Proceedings of the IEEE/CVF Conference on Computer Vision and Pattern Recognition},
  note   = {arXiv:2406.04264},
  eprint = {2406.04264},
  archivePrefix = {arXiv},
  year   = {2025}
}

@inproceedings{yu2025vrbench,
  title  = {VRBench: A Benchmark for Multi-Step Reasoning in Long Narrative Videos},
  author = {Jiashuo Yu and Yue Wu and Meng Chu and Zhifei Ren and Zizheng Huang and Pei Chu and Ruijie Zhang and Yinan He and Qirui Li and Songze Li and Zhenxiang Li and Zhongying Tu and Conghui He and Yu Qiao and Yali Wang and Yi Wang and Limin Wang},
  booktitle = {ICCV},
  year   = {2025}
}

@inproceedings{wang2024lvbench,
  title  = {LVBench: An Extreme Long Video Understanding Benchmark},
  author = {Weihan Wang and Zehai He and Wenyi Hong and Yean Cheng and Xiaohan Zhang and Ji Qi and Shiyu Huang and Bin Xu and Yuxiao Dong and Ming Ding and Jie Tang},
  booktitle = {Proceedings of the IEEE/CVF International Conference on Computer Vision},
  note   = {arXiv:2406.08035},
  year   = {2025}
}

@inproceedings{zhang2024generative,
  title  = {Generative Verifiers: Reward Modeling as Next-Token Prediction},
  author = {Lunjun Zhang and Arian Hosseini and Hritik Bansal and Mehran Kazemi and Aviral Kumar and Rishabh Agarwal},
  booktitle = {International Conference on Learning Representations},
  note   = {arXiv:2408.15240},
  year   = {2025}
}

@article{mohamed2026cost,
  title  = {A Cost-Aware, Paired Protocol for Auditing Dynamic Tool Synthesis in Agentic Video Question Answering},
  author = {Aseel Mohamed and Rama AlHamidi and Mohamed Rayan Barhdadi and others},
  journal= {arXiv preprint arXiv:2607.01469},
  year   = {2026}
}

@article{guo2026do,
  title  = {Do Multimodal Agents Really Benefit from Tool Use? A Systematic Study of Capability Gains},
  author = {Garvin Guo and Donglei Yu and Yu Chen and Xiang Wang and Shuai Li and Xinpei Zhao},
  journal= {arXiv preprint arXiv:2606.02357},
  year   = {2026}
}

@article{janjua2026perception,
  title  = {Don't Show Pixels, Show Cues: Unlocking Visual Tool Reasoning in Language Models via Perception Programs},
  author = {Muhammad Kamran Janjua and Hugo Silva and Di Niu and Bahador Rashidi},
  journal= {arXiv preprint arXiv:2604.12896},
  eprint = {2604.12896},
  archivePrefix = {arXiv},
  year   = {2026}
}

@article{he2026vtagent,
  title  = {VTAgent: Agentic Keyframe Anchoring for Evidence-Aware Video TextVQA},
  author = {Haibin He and Maoyuan Ye and Jing Zhang and others},
  journal= {arXiv preprint arXiv:2605.04870},
  year   = {2026}
}

@article{zou2026videobrain,
  title  = {VideoBrain: Learning Adaptive Frame Sampling for Long Video Understanding},
  author = {Junbo Zou and Ziheng Huang and Shengjie Zhang and others},
  journal= {arXiv preprint arXiv:2602.04094},
  year   = {2026}
}

@inproceedings{zhang2025vtimecot,
  title  = {VTimeCoT: Thinking by Drawing for Video Temporal Grounding and Reasoning},
  author = {Zhang and Guo and Potamias and Deng and Xu and Ma},
  booktitle = {Proceedings of the IEEE/CVF International Conference on Computer Vision},
  note   = {arXiv:2510.14672},
  year   = {2025}
}

@inproceedings{zheng2023judging,
  title  = {Judging LLM-as-a-Judge with MT-Bench and Chatbot Arena},
  author = {Lianmin Zheng and Wei-Lin Chiang and Ying Sheng and Siyuan Zhuang and Zhanghao Wu and Yonghao Zhuang and others},
  booktitle = {Advances in Neural Information Processing Systems 36 (Datasets and Benchmarks Track)},
  note   = {arXiv:2306.05685},
  year   = {2023}
}

@article{cobbe2021training,
  title  = {Training Verifiers to Solve Math Word Problems},
  author = {Karl Cobbe and Vineet Kosaraju and Mohammad Bavarian and Mark Chen and Heewoo Jun and Lukasz Kaiser and others},
  journal= {arXiv preprint arXiv:2110.14168},
  year   = {2021}
}

@article{verga2024replacing,
  title  = {Replacing Judges with Juries: Evaluating LLM Generations with a Panel of Diverse Models},
  author = {Pat Verga and Sebastian Hofstatter and Sophia Althammer and Yixuan Su and Aleksandra Piktus and Arkady Arkhangorodsky and others},
  journal= {arXiv preprint arXiv:2404.18796},
  year   = {2024}
}

@inproceedings{panickssery2024llm,
  title  = {LLM Evaluators Recognize and Favor Their Own Generations},
  author = {Arjun Panickssery and Samuel R. Bowman and Shi Feng},
  booktitle = {Advances in Neural Information Processing Systems 37},
  note   = {arXiv:2404.13076},
  year   = {2024}
}

@article{saluru2026disagreement,
  title  = {Disagreement-Based Cross-Model Routing for Implicit Video Question Answering},
  author = {Durga Sandeep Saluru},
  journal= {arXiv preprint arXiv:2606.14723},
  year   = {2026}
}

@article{wang2026seeing,
  title  = {Seeing Before Agreeing: Aligning Multi-Agent Consensus with Visual Evidence},
  author = {Yuhan Wang and Shuochen Chang and Yalin Feng and Dongsheng Ma and Yuanzi Li and Zhengren Wang},
  journal= {arXiv preprint arXiv:2605.30698},
  year   = {2026}
}

@article{wang2026beacon,
  title  = {Beacon: Knowing When and How to Perform Agentic Visual Reasoning},
  author = {Qixun Wang and Yang Shi and Letian Cheng and Zhuoran Zhang and Yan He and Yuqi Tang},
  journal= {arXiv preprint arXiv:2607.28595},
  year   = {2026}
}

@article{lee2026accuracy,
  title  = {Accuracy Without Grounding: Diagnosing Visual Dependency Dissociation in Video LLM Benchmarks},
  author = {Jae Joong Lee},
  journal= {arXiv preprint arXiv:2607.13305},
  year   = {2026}
}

@article{alavi2026perception,
  title  = {Perception First: A Frontier Native-Video Model with Self-Consistency for Implicit Video Question Answering},
  author = {Ali Alavi},
  journal= {arXiv preprint arXiv:2606.01485},
  year   = {2026}
}

@article{chen2026matryoshka,
  title  = {One Ranking, Any Budget: Matryoshka Evidence-to-Context Frame Selection for Long-Video Understanding},
  author = {Wang Chen and Yu Chen and Xiang Wang and Shuai Li and Jinfa Huang and Xiawu Zheng},
  journal= {arXiv preprint arXiv:2608.05707},
  year   = {2026}
}

@article{wang2026efficient,
  title  = {Efficient Frame Selection for Long Videos at Test Time with Attention-Based MLLM Selectors},
  author = {Yilin Wang and Xiangxi Zheng and Dongxing Mao and Linjie Li and Zhengyuan Yang and Ping Yu},
  journal= {arXiv preprint arXiv:2607.15689},
  year   = {2026}
}

@article{cheng2025videoholmes,
  title  = {Video-Holmes: Can MLLM Think Like Holmes for Complex Video Reasoning?},
  author = {Junhao Cheng and Yuying Ge and Teng Wang and Yixiao Ge and Jing Liao and Ying Shan},
  journal= {arXiv preprint arXiv:2505.21374},
  year   = {2025}
}

@article{zheng2024mcselectors,
  title  = {Large Language Models Are Not Robust Multiple Choice Selectors},
  author = {Zheng, Chujie and Zhou, Hao and Meng, Fandong and Zhou, Jie and Huang, Minlie},
  journal = {International Conference on Learning Representations (ICLR)},
  year   = {2024},
  eprint = {2309.03882},
  archivePrefix = {arXiv},
  primaryClass  = {cs.CL}
}

@inproceedings{pezeshkpour2023options,
  title  = {Large Language Models Sensitivity to The Order of Options in Multiple-Choice Questions},
  author = {Pezeshkpour, Pouya and Hruschka, Estevam},
  year   = {2024},
  booktitle = {Findings of the Association for Computational Linguistics: NAACL 2024},
  note   = {arXiv:2308.11483},
  eprint = {2308.11483},
  archivePrefix = {arXiv},
  primaryClass  = {cs.CL}
}

@inproceedings{wang2024fairevaluators,
  title  = {Large Language Models are not Fair Evaluators},
  author = {Wang, Peiyi and Li, Lei and Chen, Liang and Cai, Zefan and Zhu, Dawei and Lin, Binghuai and Cao, Yunbo and Liu, Qi and Liu, Tianyu and Sui, Zhifang},
  year   = {2024},
  booktitle = {Proceedings of the 62nd Annual Meeting of the Association for Computational Linguistics (Volume 1: Long Papers)},
  note   = {arXiv:2305.17926},
  eprint = {2305.17926},
  archivePrefix = {arXiv},
  primaryClass  = {cs.CL}
}

@inproceedings{shi2024judgingjudges,
  title  = {Judging the Judges: A Systematic Study of Position Bias in {LLM-as-a-Judge}},
  author = {Shi, Lin and Ma, Chiyu and Liang, Wenhua and Diao, Xingjian and Ma, Weicheng and Vosoughi, Soroush},
  year   = {2025},
  booktitle = {Proceedings of the 14th International Joint Conference on Natural Language Processing and the 4th Conference of the Asia-Pacific Chapter of the Association for Computational Linguistics},
  note   = {arXiv:2406.07791},
  eprint = {2406.07791},
  archivePrefix = {arXiv},
  primaryClass  = {cs.CL}
}

@inproceedings{loginova2024blindguessing,
  title  = {Addressing Blind Guessing: Calibration of Selection Bias in Multiple-Choice Question Answering by Video Language Models},
  author = {Loginova, Olga and Bezrukov, Oleksandr and Shekhar, Ravi and Kravets, Alexey},
  year   = {2025},
  booktitle = {Proceedings of the 63rd Annual Meeting of the Association for Computational Linguistics (Volume 1: Long Papers)},
  note   = {arXiv:2410.14248},
  eprint = {2410.14248},
  archivePrefix = {arXiv},
  primaryClass  = {cs.CV}
}

@article{tan2024ordermatters,
  title  = {Order Matters: Exploring Order Sensitivity in Multimodal Large Language Models},
  author = {Tan, Zhijie and Chu, Xu and Li, Weiping and Mo, Tong},
  year   = {2024},
  journal = {arXiv preprint arXiv:2410.16983},
  eprint = {2410.16983},
  archivePrefix = {arXiv},
  primaryClass  = {cs.CV}
}

@inproceedings{atabuzzaman2025lvlmselection,
  title  = {Benchmarking and Mitigating {MCQA} Selection Bias of Large Vision-Language Models},
  author = {Atabuzzaman, Md. and Asgarov, Ali and Thomas, Chris},
  year   = {2025},
  booktitle = {Proceedings of the 2025 Conference on Empirical Methods in Natural Language Processing},
  note   = {arXiv:2509.16805},
  eprint = {2509.16805},
  archivePrefix = {arXiv},
  primaryClass  = {cs.CV}
}

@article{wang2026morethinking,
  title  = {More Thinking, More Bias: Length-Driven Position Bias in Reasoning Models},
  author = {Wang, Xiao},
  year   = {2026},
  journal = {arXiv preprint arXiv:2605.06672},
  eprint = {2605.06672},
  archivePrefix = {arXiv},
  primaryClass  = {cs.CL}
}

@article{xu2026pointwisepairwise,
  title  = {Am {I} More Pointwise or Pairwise? Revealing Position Bias in Rubric-Based {LLM-as-a-Judge}},
  author = {Xu, Yuzheng and Hirasawa, Tosho and Kozuno, Tadashi and Ushiku, Yoshitaka},
  year   = {2026},
  journal = {arXiv preprint arXiv:2602.02219},
  eprint = {2602.02219},
  archivePrefix = {arXiv},
  primaryClass  = {cs.CL}
}

@article{norman2026reliability,
  title  = {Reliability without Validity: A Systematic, Large-Scale Evaluation of {LLM}-as-a-Judge Models Across Agreement, Consistency, and Bias},
  author = {Norman, Justin D. and Rivera, Michael U. and Hughes, D. Alex},
  year   = {2026},
  journal = {arXiv preprint arXiv:2606.19544},
  eprint = {2606.19544},
  archivePrefix = {arXiv},
  primaryClass  = {cs.CL}
}

@article{tamba2026ceiling,
  title  = {Position Bias is Hidden Behind Ceiling Effects: A Permutation Diagnostic for {LLM} Benchmarks},
  author = {Tamba, Hiroki},
  year   = {2026},
  journal = {arXiv preprint arXiv:2607.20864},
  eprint = {2607.20864},
  archivePrefix = {arXiv},
  primaryClass  = {cs.CL}
}

@inproceedings{robinson2023mcsb,
  title  = {Leveraging Large Language Models for Multiple Choice Question Answering},
  author = {Robinson, Joshua and Rytting, Christopher Michael and Wingate, David},
  year   = {2023},
  booktitle = {International Conference on Learning Representations},
  note   = {arXiv:2210.12353},
  eprint = {2210.12353},
  archivePrefix = {arXiv},
  primaryClass  = {cs.CL}
}

@misc{metaai2026musespark,
  title  = {Introducing {M}use {S}park 1.1},
  author = {{Meta AI}},
  year   = {2026},
  month  = jul,
  howpublished = {\url{https://ai.meta.com/blog/introducing-muse-spark-meta-model-api/}},
  note   = {Evaluation report: \url{https://ai.meta.com/static-resource/muse-spark-1-1-evaluation-report/}}
}

@misc{qwen2026qwen35,
  title  = {{Qwen3.5-27B}},
  author = {{Qwen Team}},
  year   = {2026},
  month  = feb,
  howpublished = {Hugging Face model card, \url{https://huggingface.co/Qwen/Qwen3.5-27B}}
}

@misc{qwen2026qwen35moe,
  title  = {{Qwen3.5-397B-A17B}},
  author = {{Qwen Team}},
  year   = {2026},
  month  = feb,
  howpublished = {Hugging Face model card, \url{https://huggingface.co/Qwen/Qwen3.5-397B-A17B}}
}

@article{abdin2024phi3,
  title  = {{Phi-3} Technical Report: A Highly Capable Language Model Locally on Your Phone},
  author = {Marah Abdin and Jyoti Aneja and Hany Awadalla and Ahmed Awadallah and Ammar Ahmad Awan and Nguyen Bach and others},
  journal= {arXiv preprint arXiv:2404.14219},
  year   = {2024},
  note   = {Introduces the phi-3.5 series, including phi-3.5-Vision}
}

@article{zhu2025internvl3,
  title  = {{InternVL3}: Exploring Advanced Training and Test-Time Recipes for Open-Source Multimodal Models},
  author = {Jinguo Zhu and Weiyun Wang and Zhe Chen and Zhaoyang Liu and Shenglong Ye and Lixin Gu and others},
  journal= {arXiv preprint arXiv:2504.10479},
  year   = {2025}
}

@article{openai2025gpt5card,
  title  = {{OpenAI GPT-5} System Card},
  author = {{OpenAI}},
  journal= {arXiv preprint arXiv:2601.03267},
  year   = {2025}
}

\appendix

\section{Limitations}
\label{appendix:limitations}

\textbf{(1)} Three benchmarks carry both deployed verifiers, over a
compressed $\approx6$--$14$-point headroom range, so our headroom result is \emph{no evidence for
the rule} rather than a correlation in either direction. \textbf{(2)} Both strong verifiers are
closed API models, though an open verifier does beat self-consistency on
EgoSchema~(Appendix~\ref{appendix:extended}). \textbf{(3)} We select over final answer letters, not
intermediate reasoning steps, and over uniformly sampled rather than query-conditioned
frames~\citep{chen2026matryoshka}. \textbf{(4)} Every benchmark we use labels its
options alphabetically, so an option's ordinal position and its letter rank are collinear; our
position-bias statistics establish the drift but, as we report them, cannot separate the two
(Appendix~\ref{appendix:extended}). No caveat here touches the oracle-level tool null of
Appendix~\ref{appendix:control}.

\section{Implementation and scoring integrity}
\label{appendix:integrity}

Beyond the models under study, the experimental pipeline itself (code, job
orchestration, and result analysis) was implemented with the assistance of LLM-based
autonomous agents under author supervision. To guard against errors from this
process, every headline number in the paper was recomputed by two independently
implemented scorers, which agree to within $0.1$ points under a fixed
convention that excludes empty trajectories from the vote. Where the harness retained the
verifier's full response the recomputation runs from it; the earlier verifier runs recorded
the extracted answer and the gold label but not the response behind them, so for those legs
the answer-extraction step cannot be re-audited after the fact. Where a quantity was
re-measured at a larger generation compute budget, the re-run reproduces the original run's per-row
answers on $99.4$--$99.9\%$ of rows when its output is truncated back to the original
compute budget; the same-family self-verifier on MLVU-Test, at $95.2\%$, is the lone exception. On
the positive-control verifier every disagreeing row is one whose response was shorter than
the original cap, which attributes that residual to run-to-run generation nondeterminism
rather than to the cap. Prediction
and pick files were also checked directly rather than through logged summaries: gold
labels were cross-verified, duplicates removed, and empty or malformed outputs
handled explicitly.

\section{Protocol and denominators}
\label{appendix:protocol}

This appendix records in full the conventions that Section~\ref{section:setup} states in
summary: the majority-vote tie-break, the three scoring policies and their out-of-candidate
rates, the split sizes, and the per-cell denominators behind the anchor table. Every figure
here is as measured.

\paragraph{Splits.} Unless noted, results use each benchmark's full test split
($\approx500$--$1{,}850$ items). VRBench is the one exception: its cross-model
(Muse / GPT-5.6 Sol) cells use a $1{,}000$-question subset, while its Phi-3.5 and
InternVL3 cells use a larger $\approx1{,}500$-item split. We flag the subset at each
VRBench table and figure, and never compare across the two VRBench splits---every
VRBench delta is computed within one of them.

\paragraph{Tie-breaking.} Ties in the majority vote are broken by \emph{first appearance}---the option whose earliest supporting chain has the lowest index---which we verified directly against the shipped implementation on all $1{,}837$ Video-Holmes ballots at $N{=}16$: the shipped choice matches first appearance on every row, including all $49$ rows on which the vote is tied and the $26$ on which an alphabetical rule would have chosen differently, so an alphabetical tie-break is falsified outright. The rule is also effectively letter-neutral. Against a letter-blind tie-break---one that resolves a tie uniformly at random among the tied options---it never moves the A-share by more than $0.6$ points at any $N$ on any benchmark in our suite; the largest departure we measure is $0.57$ points, on EgoSchema at $N{=}8$.

\paragraph{Sensitivity of the majority deficit to the tie-break rule.} Because the tie-break is
an artifact of sampling order rather than a principled rule, we bounded its influence directly.
Ties are not rare---they occur on $3.98\%$ of MLVU-Test rows and $6.26\%$ of Video-Holmes rows (empty
ballots excluded throughout, per the convention of Section~\ref{section:setup}),
and the gold answer sits inside the tied set on well over half of them---so we recomputed
$\Delta_{\text{maj}}$ for all six Phi-3.5 cells under three gold-blind rules (first appearance;
alphabetically first among the tied options; and alphabetically last), under both
scored fields. All $36$ resulting comparisons are negative and significant at $\alpha{=}0.05$,
the loosest at $p{=}8.1\times10^{-3}$, and all $36$ pass the count identity of
Section~\ref{section:setup}: neither the direction nor the significance of the weak verifier's
majority deficit depends on the rule. The \emph{magnitude} does. The gold-blind spread reaches
$1.23$ points on MLVU-Test, and an envelope that additionally admits gold-aware bounds reaches
$4.30$ points on Video-Holmes, so we do not rank cells by $\Delta_{\text{maj}}$ at a resolution
finer than about $1.2$ points, and no claim in this paper rests on an exact $\Delta_{\text{maj}}$
magnitude. The position-bias grid is untouched by this seam: A-lift compares the selector's pick
against gold and never consults the majority vote at all.

\paragraph{The flagship family, enumerated.} The three-cell family over which we apply
Holm--Bonferroni is the set of three $\Delta_{\text{maj}}$ comparisons on Video-Holmes made by
the two deployed cross-model verifiers: GPT-5.6 Sol on the Qwen-base pool
($p{=}7.0\times10^{-4}$), Muse on the Qwen-base pool ($p{=}2.3\times10^{-3}$), and Muse on the
Muse-base pool ($p{=}0.331$). The family-wise level is $\alpha{=}0.05$. Both Qwen-base cells
survive correction. The Muse-base cell survives at no family size and does not reach
significance uncorrected, so it is a null we report rather than a survivor that drops out.
Enlarging the family---for instance to the five cells the Solver policy makes available---leaves
both survivors intact, but we do not offer that as independent confirmation, since it
re-tests the same two comparisons against a larger family drawn from the same data. What
the enlargement does establish is a robustness bound: under Holm, adding a comparison
tightens an incumbent's threshold by one rung whenever the added $p$ is at least the
incumbent's own, so the worst case for an incumbent at family size $M$ is the Bonferroni
level $\alpha/M$. Both survivors clear that level at every family size up to $M{=}21$
(Sol up to $M{=}70$), so no plausible enumeration of the comparison family overturns them.

\paragraph{Scoring policies.} A verifier asked to select can solve instead: it may
emit a letter outside the $N$-candidate set, and if that out-of-candidate (OOC)
letter happens to be correct, naive scoring credits it as a successful selection.
OOC rates vary by benchmark and verifier. Across the six deployed cross-model cells they are
$4.70\%$ and $4.84\%$ on TempCompass, $5.04\%$ and $8.96\%$ on MLVU-Test, and $1.73\%$ and
$0.71\%$ on VRBench (Muse-48 and GPT-5.6 Sol respectively); the deployed Video-Holmes cells
span $0.9$--$6.1\%$; and the highest rate we observe anywhere is $\approx13\%$, on a weak
verifier. Verifier strength alone does not order them: the stronger verifier's rate is nearly
twice the weaker one's on MLVU-Test and under half of it on VRBench. We therefore define three
scoring policies. \emph{Solver} counts OOC
hits (the default in prior work). \emph{Selector} counts a pick only if it is
correct and belongs to the sampled candidate set; a correct pick outside the set is
scored wrong, since it reflects solving rather than selection. \emph{Snap} maps an
OOC pick to the nearest candidate. Headline numbers below use the \emph{Solver} policy,
the default in prior work, so our accuracies are directly comparable to published
best-of-$N$ results; we report the stricter Selector policy as a clearly-labeled
ablation and flag where the two disagree. The gap between Solver and Selector scoring is exactly the leakage
a headline number would over-credit: for the cross-model verifier, whose out-of-candidate rate stays under $9\%$ on every deployed split, Solver scoring still inflates its
apparent selection accuracy. Under Selector scoring the same six cells read $83.3\%$ and $84.3\%$ on TempCompass, $72.8\%$ and $69.1\%$ on MLVU-Test, and $87.7\%$ and $89.3\%$ on VRBench (Muse-48 and GPT-5.6 Sol respectively), against the Solver accuracies of Table~\ref{table:headline}; the ablation moves the headline by $3.88$, $4.27$, $1.21$, $1.79$, $0.10$ and $0.00$ points in that order. The out-of-candidate correction applies only to \emph{selection}: the standalone solver never sees $\mathcal{C}$, so the standalone leg of the Selection Increment carries no out-of-candidate adjustment and is identical under all three policies.

\paragraph{Statistics.} Every headline delta is matched-sample and tested with an exact two-sided paired McNemar test on
the discordant pairs. Because we evaluate a grid of verifiers
and benchmarks, we report per-comparison McNemar significance, flag non-significant
deltas (n.s.) throughout, and treat isolated significant results cautiously given the
number of comparisons. Implementation and scoring-integrity details are in
Appendix~\ref{appendix:integrity}.

\paragraph{Denominators on the anchor.} Majority@8, pass@1 and oracle@8 are computed on all $1{,}837$
Video-Holmes questions. A chain that yields no extractable option letter ($1.06\%$ of the
$14{,}696$ chains; no question loses all $N$) casts no ballot and is excluded from the
pass@1 average, which we take per question and then average over questions, matching the majority@8 convention. The selector cells are computed on the items for which the verifier
returned a parseable pick: $1{,}834$ for GPT-5.6 Sol and $1{,}825$/$1{,}824$ for the Muse
verifier on the Qwen-/Muse-generated pools, following the matched-sample convention of
Section~\ref{section:setup}. The shortfall is entirely verifier-side: three video-decode
failures inside the GPT verifier, and $12$/$13$ Muse responses that completed but contained no
extractable option letter. No item is lost at load time---every pick file contains all $1{,}837$
question keys, with no duplicates and with gold labels identical to the released annotations, and
a zero-shot $N{=}16$ run on the same split scores all $1{,}837$. Scoring the unanswered items as
incorrect rather than excluding them leaves our conclusions unchanged: GPT-5.6 Sol
$+3.11\rightarrow+2.99$ points over self-consistency (McNemar $p{=}7.0\times10^{-4}\rightarrow
1.1\times10^{-3}$) and the Muse verifier on the Qwen-base pool $+2.85\rightarrow+2.40$
($p{=}2.3\times10^{-3}\rightarrow1.1\times10^{-2}$), both still significant under the Solver policy after
Holm correction across the three-cell family. Under the stricter Selector ablation it does move
the GPT-5.6 Sol margin from $+1.80$ ($p{=}0.043$) to $+1.69$ ($p{=}0.059$); we report both and
treat that margin as not established, consistent with its failure to survive Holm correction
under either denominator.

\section{Negative control: gathering more evidence does not raise the ceiling}
\label{appendix:control}

\textbf{On-demand tools add no selectable benefit, even at the oracle.} A natural objection is
that selection looks limited only because the sampled candidates are impoverished, and that
letting the model \emph{gather} more evidence would raise the ceiling a verifier works
against. It does not. Augmenting \emph{candidate generation} with an on-demand agentic loop
that may zoom, crop, retrieve frames, or ground objects leaves the best-of-$N$ oracle on EgoSchema unchanged ($\Delta_{\text{oracle}}{=}0.000$ at $n{=}500$). The reason is
mechanical: selection can only surface a correct answer the sampler already produced, and the
tools produce no new correct trajectory to surface, so even a hypothetical error-free verifier
would gain nothing from them. Nor is the null a weak-model artifact---giving the same evidence
loop to a strong cross-model verifier (Muse) leaves its accuracy within noise of its
zero-shot self ($-2.60$ points at $n{=}500$, a small, non-significant change). Tool-augmented
inference does not move the ceiling that selection works against on either side of the
pipeline, which is what directs this study at selection rather than gathering.

\section{Extended results}
\label{appendix:extended}

\subsection{Response-mode decomposition of the position bias}
\label{appendix:modes}

Table~\ref{table:modes} gives the decomposition behind
Section~\ref{section:position}. The weak verifier's responses partition into two modes that
output length separates without overlap---five tokens for the marker form
\texttt{Answer:\,X}, three for a bare letter---and every response in both modes
terminates normally, with no truncation at the token
budget.\footnote{One response in $9{,}021$ emits a letter outside the option set; it yields no
pick and enters no cell here. Normal termination and the absence of budget truncation hold on
all $9{,}021$.} The
partition is therefore a property of the verifier's output, not of our parser: the strict answer-marker
parse is inert on this subset, and holding
the row set fixed while swapping only the scored field changes no cell. What separates the
two mode columns below is which rows are present, not how they are scored.

The first-option preference is confined to the terse mode in all eight arms in which that
mode occurs, and no marker-mode arm reaches significance in any of the ten arms measured---a
null we read as not distinguishable from zero at these per-arm $n$, the closest cell missing
$\alpha$ at $p{=}0.0501$. The
split is not a restatement of accuracy: the terse mode is less accurate in seven of the eight
arms, but on EgoSchema at $K{=}16$ the two modes are accurate to within $0.11$ points and the
terse mode is still significantly biased, so the preference is a directional default rather
than a by-product of solving worse.

That makes the mode contrast confounded with the item at the arm level, so we also test it
within items. Restricting to the items whose mode changes across frame counts holds the
question, its options, its gold answer and its difficulty fixed by construction, though not
the frame count itself: the terse and marker observations of such an item necessarily come
from different frame budgets. Terse observations nevertheless favor the first option by
twenty-five points more than marker observations of the same items on MLVU-Test, and by
eighteen points more on Video-Holmes (exact two-sided sign tests, $p{=}4.8\times10^{-7}$ and
$1.9\times10^{-17}$; zero and four of $81$ and $340$ items respectively run the other way).
The frame budget does not explain this. The contrast is positive in both switch directions and
in every frame-count pair taken separately, so it cannot be a monotone effect of frame count;
on the items whose terse and marker observations differ by only four frames it is larger, not
smaller ($+29.5$ and $+38.8$ points); and the frame effect measured on items whose mode never
changes---where the frame budget varies but the mode does not---is at most $2.4$ points, an
order of magnitude too small to produce it.

The capable InternVL3-38B control emits the marker on all
$502$ items of both its arms, produces no terse response at all, and reaches significance in
neither ($p{=}0.098$ and $p{=}0.161$), which is the outcome we predicted before measuring it.
Because that arm carries no terse rows, however, mode and model strength are collinear within
it: it discharges extraction failure and item difficulty as explanations, but it cannot
separate the response mode from verifier strength, and we do not read it as establishing that
the strong verifier is unbiased.

What does separate mode from item is an item-matched stratification. Scoring InternVL3-38B on
exactly the items where the weak verifier was terse---the same $502$-item pool, an identical
gold map---leaves it at $+1.02$ points on the $197$ items of the $K{=}24$ arm ($p{=}0.85$) and
at $0.00$ on the $227$ items of the $K{=}48$ arm ($p{=}1.000$), against the $+27.9$ and
$+29.1$ that the weak verifier posts on those very items
(Table~\ref{table:modes}). The terse rows are therefore not merely A-heavy items. Two scoping
notes attach to this. Under the same stratification InternVL3-38B's marker cells do move to
significance, in the direction of a \emph{deficit} at the first option rather than an excess,
so the tally we report scores each arm on its own full $502$-item cell. And the two
InternVL3-38B arms are near-duplicates---they agree on $486$ of $502$ picks---so they furnish
nine effectively independent marker tests rather than ten.

\begin{table}[t]
\centering\small
\caption{Position bias by response mode, weak verifier (Phi-3.5-Vision). Modes are
disjoint and separated by output length. All tests are exact two-sided binomial McNemar on
non-empty picks; A-lift is
$100(b_A-c_A)/n$. Marker-mode A-lift magnitudes are withheld pending independent
verification: no marker-mode cell reaches significance in any of the ten arms as
scored here, each on its own full row set, and every marker-mode $|\text{A-lift}|$ is below
$4$ points, though at these per-arm $n$ that is a
failure to distinguish the marker mode from zero rather than a demonstration that it is
unbiased---the closest cell sits just above $\alpha$, at $p{=}0.0501$. Accuracies are under the Solver policy.}
\label{table:modes}
\begin{tabular}{lrrrrrrr}
\toprule
& \multicolumn{3}{c}{Marker mode (\texttt{Answer:\,X})} & \multicolumn{4}{c}{Terse mode (bare letter)} \\
\cmidrule(lr){2-4}\cmidrule(lr){5-8}
Verifier--benchmark arm & $n$ & acc. & $p$ & $n$ & acc. & A-lift & $p$ \\
\midrule
MLVU-Test, $K{=}24$ & $305$ & $71.5$ & $0.0501$ & $197$ & $54.8$ & $+27.9$ & $1.5\times10^{-12}$ \\
MLVU-Test, $K{=}48$ & $275$ & $70.2$ & $0.720$ & $227$ & $56.8$ & $+29.1$ & $1.3\times10^{-16}$ \\
MLVU-Test, $K{=}64$ & $326$ & $69.3$ & $0.311$ & $176$ & $54.0$ & $+32.4$ & $9.0\times10^{-15}$ \\
Video-Holmes, $K{=}24$ & $1{,}241$ & $53.8$ & $0.729$ & $596$ & $47.0$ & $+13.4$ & $2.3\times10^{-10}$ \\
Video-Holmes, $K{=}48$ & $1{,}225$ & $53.4$ & $0.346$ & $612$ & $41.7$ & $+24.0$ & $6.8\times10^{-25}$ \\
Video-Holmes, $K{=}64$ & $1{,}432$ & $52.3$ & $1.000$ & $404$ & $38.6$ & $+26.7$ & $5.4\times10^{-19}$ \\
EgoSchema, $K{=}16$ & $185$ & $69.7$ & $1.000$ & $315$ & $69.8$ & $+6.7$ & $2.5\times10^{-3}$ \\
EgoSchema, $K{=}48$ & $232$ & $67.7$ & $0.701$ & $268$ & $63.4$ & $+13.8$ & $7.5\times10^{-7}$ \\
\bottomrule
\end{tabular}
\end{table}

\paragraph{Why the filtered view hides a real bias.} The answer-marker partition acts entirely
on the denominator, and not neutrally. It removes $22$--$63\%$ of responses, and the removed
rows are the biased ones: the rate at which the verifier picks ``A'' differs sharply between
retained and removed rows (Fisher exact $p<10^{-13}$ in all six MLVU-Test and Video-Holmes
cells, as low as $1.3\times10^{-35}$; $p{=}2.4\times10^{-2}$ and $6.7\times10^{-4}$ on the two
EgoSchema arms), while the rate at which ``A'' is \emph{gold} is matched between them (exact
two-sided Fisher, sum-of-tables convention, $p{=}0.30$ to $0.74$ across all eight arms). The
removed rows are therefore not differently keyed. Neither are they materially harder: on a
verifier-blind difficulty proxy---the majority@8 accuracy of the same generator pool---the two
sets are statistically indistinguishable on Video-Holmes and EgoSchema ($p{=}0.13$ to $1.00$),
and the one gap that does appear, on MLVU-Test ($66.5\%$ versus $77.0\%$ at $K{=}24$,
$p{=}0.010$ uncorrected, $0.062$ after Holm correction over the six cells), is about half a
task-type composition effect: the terse rows over-sample MLVU-Test's hardest task types, and
within task type the gap narrows to $+5.1$ points and is not significant ($p{=}0.17$ to
$0.26$). What separates the two sets is overwhelmingly the verifier's own letter
preference---they are the rows on which this verifier misbehaves. Conditioning on the marker is thus a selection on the verifier's own
behavior, which is why a preference that is real in the full sample is not visible in the
filtered one.

\subsection{Provenance of the verifier ladders}\label{appendix:provenance}
(i) \emph{The two ladders are not independent}: the Qwen-27B self-verifier rung appears on both, measured under a $1{,}024$-token cap that bound its solving far harder than its picking, so its standalone is a lower bound and its increment an upper bound on both. Re-measured at $8{,}192$ tokens it moves toward the strong-solver end of each ladder and its increment falls with it---each ordering survives, but MLVU-Test is not evidence independent of Video-Holmes. (ii) \emph{Phi-3.5 runs at $24$ frames}, against $48$ for every other rung. Both its legs share that compute budget, so its SI is a matched difference; but a ladder sweeps standalone accuracy, so part of why this rung solves poorly is protocol, not capability. (iii) \emph{The GPT-5.6 Sol rung is differenced on its own matched non-empty support}: three of its selector ballots and one of its standalone ballots are empty, and the two sets are disjoint, so that rung's increment rests on $n{=}1{,}833$ rather than the full $1{,}837$ split. Scoring those four ballots wrong and differencing on all $1{,}837$ instead moves the rung to $+7.7$ from a $54.2\%$ standalone.

\subsection{The verifier's frame budget, rung by rung}
The verifier's frame budget is not a lever in either direction. Across six paired comparisons on these two benchmarks at $24$, $48$ and $64$ frames, no rung improves selection and every point estimate is $\le0$; but the only significant degradations cross the $24\!\to\!48$ step, which also carries a forced doubling of the engine context length and a halving of the decode batch size, so we do not attribute them to frames alone. The single rung that isolates the frame budget under an otherwise identical engine configuration ($48\!\to\!64$) is null on both benchmarks (MLVU-Test $-0.20$ points, $p{=}1.00$; Video-Holmes $-0.16$, $p{=}0.82$).

The weak Phi-3.5 verifier's deficit against the vote also replicates on EgoSchema: $\Delta_{\text{maj}}$ is $-6.8$ points at $16$ frames (exact McNemar $p{=}7.6\times10^{-5}$) and $-11.2$ at $48$ frames ($p{=}3.1\times10^{-9}$), against a majority@8 of $76.6\%$ on the EgoSchema pool ($n{=}500$). Both cells are deficits, stated with their sign; they sit outside the Holm-corrected flagship beats-majority family and we report them as a separate replication.

This appendix collects the tables and figures supporting Section~\ref{section:rule}
and Section~\ref{section:position} that the page limit keeps out of the main text.
Every number is as measured; no cell is recomputed here.

\begin{figure}[htbp]
    \centering
    \includegraphics[width=0.82\textwidth]{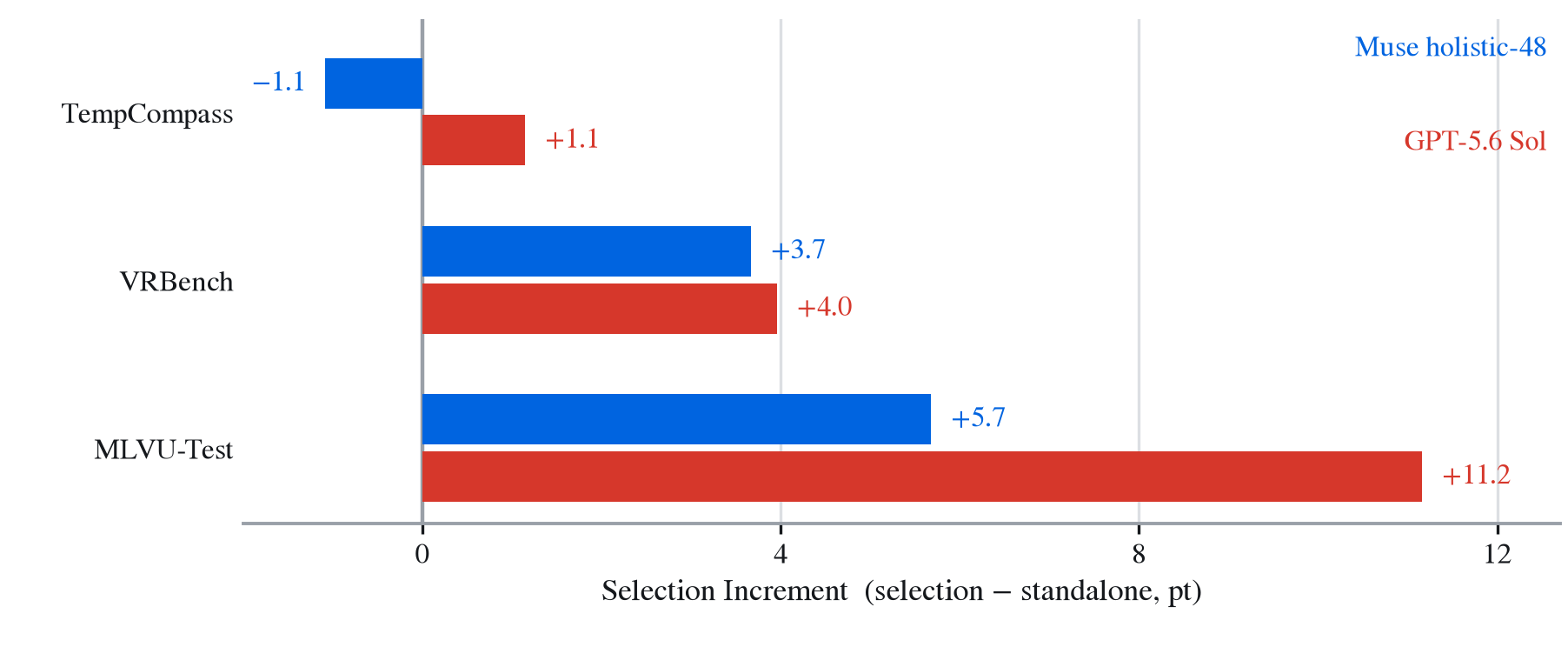}
    \caption{\textbf{Selection Increment by benchmark and verifier.} SI
    (selection accuracy minus the same verifier answering directly, both under the Solver policy) for the Muse
    holistic-48 verifier (blue) and GPT-5.6 Sol (red). GPT-5.6 Sol yields positive SI on all three benchmarks while Muse splits (negative on TempCompass)---not because GPT is
    the stronger selector but because its standalone accuracy typically sits below Muse's,
    leaving the candidates more room, consistent with SI tracking standalone accuracy
    inversely (Section~\ref{section:rule}). VRBench cells use the first 1{,}000 questions of its split ($118$ of $960$ videos).}
    \label{fig:si}
\end{figure}

\begin{table}[htbp]
    \centering
    \small
    \caption{\textbf{Base $\times$ verifier on TempCompass (full $2\times2$).}
    Selection Increment and gain over self-consistency ($\Delta_{\text{maj}}$), in points, for a
    weak (Qwen-27B) and a strong (Muse) verifier over each of two candidate pools. SI is
    a \emph{column} effect (large for the weak verifier, near zero for the strong one on
    both bases); beating self-consistency is a \emph{row} effect (attainable on the
    harder Qwen-base pool, out of reach on the near-saturated Muse-base pool). The two
    axes are governed by different factors. The three small $\Delta_{\text{maj}}$ cells ($+0.9$,
    $-0.3$, $-0.4$) are effectively at parity; the row effect is clearest in the Muse
    column ($+7.9\!\to\!-0.4$).
    \emph{Provenance.} SI is reported with both of its legs---selection and standalone---at
    the original $1{,}024$-token generation cap, since re-measuring one leg alone would mix compute budgets within a single difference. Each SI cell is differenced on its verifier's matched non-empty support. The two Qwen-27B $\Delta_{\text{maj}}$ cells are re-measured
    at $8{,}192$ tokens with a corrected answer extractor; the Muse cells never bound the cap
    but retain the original extractor (Appendix~\ref{appendix:integrity}). In both base-varying $2\times2$s the base axis moves the generator's frame budget together with its identity---the Qwen-base pools read $64$ frames, the Muse-base pools $32$---so a row effect is a property of each pool as produced rather than of model identity alone.}
    \label{table:swap2x2}
    \begin{tabular}{llrr}
    \toprule
    Base pool & Verifier & SI & $\Delta_{\text{maj}}$ \\
    \midrule
    Qwen-base ($n{\approx}1{,}579$, maj $79.2$) & Qwen-27B & $+8.7$  & $+0.9$ \\
    Qwen-base                                & Muse     & $-1.1$  & $+7.9$ \\
    Muse-base ($n{\approx}1{,}000$, maj $87.5$) & Qwen-27B & $+18.5$ & $-0.3$ \\
    Muse-base                                & Muse     & $+0.5$  & $-0.4$ \\
    \bottomrule
    \end{tabular}
\end{table}

\begin{table}[htbp]
    \centering
    \small
    \caption{\textbf{Answer-position bias on EgoSchema is a weak-verifier property.}
    Per-verifier A-lift (pick-``A'' rate minus gold-``A'' rate, in points, under the
    Solver scoring policy; gold-``A'' $=20.2\%$, gold mode E, $n{=}500$), ordered weak to
    strong. Only the weak cross-family verifier Phi-3.5-Vision lifts ``A'' above gold, at both of its compute budgets; every capable verifier stays within $1.4$ points of gold.}
    \label{table:position}
    \begin{tabular}{llc}
        \toprule
        Verifier & Class & A-lift (pp) \\
        \midrule
        Phi-3.5-Vision @16f          & weak, cross-family    & $+4.4$ \\
        Phi-3.5-Vision @48f          & weak, cross-family    & $+8.0$ \\
        \midrule
        Muse (holistic-48) & strong, cross-model   & $+1.4$ \\
        InternVL3-8B @48f            & mid, cross-family     & $+0.6$ \\
        InternVL3-38B @48f           & mid, cross-family     & $+1.4$ \\
        Qwen3.5-27B (self-verifier) @48f & same-family (self) & $+0.6$ \\
        Qwen3.5-397B @48f            & same-family (sibling) & $+0.2$ \\
        Qwen3.5-397B @96f            & same-family (sibling) & $+0.8$ \\
        \bottomrule
    \end{tabular}
\end{table}

\begin{table}[htbp]
    \centering
    \small
    \caption{\textbf{Answer-position bias is a pathology of under-capable verifiers.} Per-verifier A-lift
    (pick-``A'' rate minus gold-``A'' rate, in points; under the Solver scoring policy) across benchmarks. Where a
    benchmark induces the bias, the weak verifier (Phi-3.5) lifts ``A'' above gold, whereas the capable
    same-family self-verifier sits \emph{below} gold on both cells we re-measured without a truncating
    generation cap (MLVU-Test $-2.2$, Video-Holmes $-2.8$); no capable cross-family verifier lifts ``A''
    above gold by more than $1.4$ points (on Video-Holmes the frontier GPT-5.6 Sol is
    likewise robust, $-1.2$). TempCompass induces no default for any verifier.
    \emph{Compute budget caveat.} The Qwen3.5-27B \emph{(self)} column was first measured at a $1{,}024$-token
    cap that truncated the verifier's own response. Four cells were re-measured at $8{,}192$ tokens
    with a corrected answer extractor. Both Qwen3.5-27B cells change sign---by $-8.0$ points on
    MLVU-Test and $-5.0$ on Video-Holmes---and raising the cap alone accounts for $-6.4$ and $-4.4$
    of that, flipping both signs by itself; the extractor only deepens them. The other two are
    Phi-3.5 positive controls: its cap and extractor effects are exactly zero (its longest
    MLVU-Test response is five tokens, so the cap never bound), and re-measuring moved its
    Video-Holmes A-lift by only $0.22$ points---from the $+4.2$ printed here to $+4.0$---through
    run-to-run sampling alone. That residual, against flips of $8.0$ and $5.0$ points, is what
    attributes the reversals to the cap rather than to the act of re-measuring. Phi-3.5 entries are
    the original run; the significance tests of Section~\ref{section:position} use the re-measured
    one, which changes no survivor. Every remaining entry, including the other two Qwen3.5-27B cells ($-0.8$,
    $+0.6$), is uncorrected: original extractor, $1{,}024$-token cap.
    Phi-3.5 on EgoSchema is shown at $48$ frames; dashes mark cells not run.}
    \label{table:alift}
    \begin{tabular}{lrrrr}
    \toprule
    Benchmark & Phi-3.5 \emph{(weak)} & Qwen3.5-27B \emph{(self)} & InternVL3-8B & InternVL3-38B \\
    \midrule
    TempCompass & $0.0$  & $-0.8$ & $-1.6$ & $-2.0$ \\
    MLVU-Test   & $+8.6$ & $-2.2$ & $-3.4$ & $-2.8$ \\
    VRBench     & $+3.0$ & ---    & ---    & $-0.6$ \\
    EgoSchema    & $+8.0$ & $+0.6$ & $+0.6$ & $+1.4$ \\
    Video-Holmes & $+4.2$ & $-2.8$ & $-1.6$ & $-1.9$ \\
    \bottomrule
    \end{tabular}
\end{table}

\subsection{Verifier strength and the two axes, in full}

The paragraphs below are the long-form analysis behind Section~\ref{section:rule}.

Table~\ref{table:headline} bears this out. On the full TempCompass split ($n{=}1{,}574$) the cross-model verifier reaches $87.2\%$ against $79.2\%$ for self-consistency, a gain of $7.94$ points---its largest anywhere. The edge closes, then reverses, as the splits saturate: on MLVU-Test selection only ties self-consistency ($+0.81$, n.s., $n{=}496$), and on VRBench, whose self-consistency already reaches $91\%$, it \emph{loses} ($-3.26$, McNemar $p{=}5.6\times10^{-6}$). These accuracies use the Solver convention standard in prior work (Section~\ref{section:setup}); the candidates-only ablation, which removes any credit the verifier could earn by answering the question itself rather than selecting, moves them by $0.0$--$4.3$ points and is reported in Appendix~\ref{appendix:protocol}. On VRBench with the GPT-5.6 Sol verifier the two policies coincide exactly, because none of the seven out-of-candidate picks there is correct.

\paragraph{The ``headroom'' rule does not predict selection benefit.} A natural
hypothesis---the ``headroom'' rule---holds that selection helps in proportion to the room
the sampler leaves: the headroom $\mathrm{h}=\text{oracle}-\text{majority}$, the gap
between the oracle and the self-consistency (majority voting) baseline, with a break-even
threshold below which selection should not beat self-consistency. The data give it no
support. Across our benchmarks the verifier's gain over self-consistency is only weakly
and non-significantly correlated with headroom, and individual benchmarks violate any
monotone version in \emph{both} directions: TempCompass, at low headroom ($\approx11$
points), beats self-consistency by $+9.4$ points ($p{<}10^{-23}$), whereas MLVU-Test, at
higher headroom ($\approx14$ points), \emph{underperforms} it ($-2.0$, n.s.). VRBench, at
the lowest headroom of all ($\approx6$ points), also loses ($-1.6$ to $-3.3$,
McNemar-significant), consistent with the rule rather than against it. We report this as a
negative result and caveat it honestly: with three benchmarks spanning a compressed
$\approx6$--$14$-point range, we do not claim a reliable correlation in either direction,
only that \emph{there is no evidence for a headroom rule}. Headroom should not be confused
with the \emph{level} of the majority baseline, which does carry signal: the two are
different quantities---one a gap, the other a position---and only the second survives
below. What predicts the outcome is the verifier itself, once the baseline leaves it
anything to win.

\paragraph{The Selection Increment falls as a verifier's standalone accuracy rises.} The
cleanest read comes from holding the candidate pool fixed and swapping a weak verifier for a
strong one. On Video-Holmes the weak Qwen-27B verifier (standalone $27.7\%$) posts a large
increment (SI $+28.7$ and $+31.2$ on the two base pools), while the strong Muse verifier
(standalone $59.4\%$) posts essentially none ($+2.5$ and $+0.9$); on TempCompass the same
swap moves SI from $+8.7$ (weak Qwen) to $-1.1$ (strong Muse) on the Qwen-base pool
(Tables~\ref{table:vh2x2},~\ref{table:swap2x2}). The pattern is monotone: sweeping verifiers
of increasing standalone accuracy over a fixed pool drives SI toward zero. A five-verifier
ladder on the Video-Holmes Qwen-base pool makes this explicit---SI falls from $+28.7$
(Qwen-27B self-verifier, standalone $27.7\%$) through $+19.3$ (Phi-3.5, $32.5\%$), $+17.6$
(InternVL3-8B, $39.7\%$) and $+10.9$ (InternVL3-38B, $47.7\%$) to $+7.9$ (GPT-5.6 Sol,
$54.1\%$)---and an MLVU-Test ladder gives the same ordering ($+26.3$, $+21.6$, $+20.8$,
$+10.8$ as standalone rises $38.6\!\to\!46.7\!\to\!51.1\!\to\!62.0\%$).\footnote{Three provenance notes---the shared Qwen-27B rung, the Phi-3.5 frame budget, and the GPT-5.6 Sol support---are stated in full in Appendix~\ref{appendix:provenance}.} A large SI therefore
marks a weak \emph{solver} that can nonetheless select, not a strong verifier: SI is an increment net of the verifier's own solving---how much the candidates add to a given verifier---not a capability score.

Read this way, the one swap that appears to run the other direction is not a contradiction.
Replacing Muse with the closed GPT-5.6 Sol raises SI on every deployed benchmark---to
$+1.15$ from $-1.08$ (TempCompass), $+11.16$ from
$+5.68$ (MLVU-Test) and $+3.96$ from $+3.67$ (VRBench) (Table~\ref{table:headline},
Figure~\ref{fig:si})---not because
GPT is the stronger selector but because it \emph{under-solves} these benchmarks by more than
Muse does: its standalone typically sits below Muse's (on TempCompass $87.4\%$ vs.\ $88.3\%$,
partly because it reads fewer frames), so it has more room for the candidates to help.
(Consistently, on Video-Holmes GPT is the strongest solver of the five verifiers above---and
so posts the \emph{smallest} increment there.) Both swaps thus obey one rule---SI grows as a
verifier's own standalone accuracy falls---rather than tracking raw verifier strength.

\paragraph{Video-Holmes isolates the verifier lever.} The second axis---beating
self-consistency---is cleanest on Video-Holmes, our primary reasoning anchor, because
there the two candidate pools are near-equally hard: majority vote sits at $58.9\%$ on the
Qwen-base pool and $59.6\%$ on the Muse-base pool. With base difficulty held nearly constant---though the Qwen-base pool reads $64$ frames and the Muse-base pool $32$---the verifier decides the outcome (Table~\ref{table:vh2x2}). Under the Solver policy the strong Muse verifier significantly beats majority on the Qwen-base pool ($+2.8$; McNemar $p{=}2.3\times10^{-3}$, Holm-corrected over the three-cell flagship family) and numerically edges it on the near-tied Muse-base pool ($+0.8$; $p{=}0.33$, n.s.), whereas the weak Qwen verifier fails to beat it on either ($+0.05$ and $+0.00$; exact McNemar $p{=}1.000$ on both---exact draws, not marginal nulls); base has little say. \emph{Provenance for the $2\times2$:} the two $\Delta_{\text{maj}}$ cells in the Qwen-27B column are re-measured at $8{,}192$ tokens with a corrected answer extractor; as originally measured at a $1{,}024$-token cap that truncated the verifier's own response they read $-2.5$ and $-0.7$, the second never significantly ($p{=}0.26$), and both corrected cells are exact draws. Panel~A of Table~\ref{table:vh2x2} is not re-measured, so for that column the two panels rest on different measurements of the same selection accuracy and are not additive (Appendix~\ref{appendix:integrity}). The standalone $27.7$ that Panel~A differences against was measured under that same $1{,}024$-token cap, which bound solving far harder than picking, so this column is an upper bound on SI; the paper's dissociation claim is anchored instead on the Phi-3.5 ladder rung, whose two legs share one token budget. The two-axis dissociation takes its sharpest form on this same Qwen-base pool. The weak
Phi-3.5 verifier selects far better than it solves---SI $+19.3$ from a $32.5\%$
standalone---yet its selected accuracy lands $7.1$ points \emph{below} the majority vote
($51.8\%$ vs.\ $58.9\%$; exact McNemar $p{<}10^{-13}$), whereas the strong Muse
verifier adds almost nothing over its own solving ($+2.5$) yet beats it. The
same pattern runs the length of the five-verifier ladder over this pool: all four open
verifiers post large increments and none beats the $58.9\%$ majority; only the frontier
GPT-5.6 Sol clears it ($62.0\%$, $+3.1$; McNemar $p{=}7.0\times10^{-4}$). Under the Solver policy both the Qwen-base Muse beat and this GPT
beat survive Holm--Bonferroni correction across the three-cell flagship family; under the
stricter candidates-only ablation neither does, and we report that margin as not
established. Phi-3.5's deficit is not confined to Video-Holmes: the same verifier
reproduces it on MLVU-Test---$8.0$ points below the majority vote there ($64.9\%$ vs.\
$72.9\%$, $n{=}502$; exact McNemar $p{=}2.4\times10^{-6}$)---against a different solver
pool and a different baseline, with neither cell truncating at the generation cap. Both
cells do, however, share the verifier's $24$-frame budget---a confound that replication
across benchmarks cannot discharge---so we re-ran the verifier on both at the generator's
own $64$ frames. Parity does not close either deficit; it leaves it numerically larger, at
$9.6$ points below the vote on Video-Holmes ($49.3\%$ vs.\ $58.9\%$, $n{=}1836$; exact
McNemar $p{=}3.3\times10^{-22}$) and $9.0$ on MLVU-Test ($63.9\%$ vs.\ $72.9\%$,
$n{=}502$; $p{=}1.0\times10^{-6}$). A frame handicap predicts improvement as the verifier
moves toward the generator's compute budget; we measure degradation, which falsifies that account
without establishing what the real mechanism is.

\paragraph{On TempCompass, base saturation takes over.} TempCompass shows the second gate
at work. Its full $2\times2$ (Table~\ref{table:swap2x2}) toggles a Qwen-base pool
($n{\approx}1{,}579$, majority $79.2\%$) against a stronger Muse-base pool
($n{\approx}1{,}000$, majority $87.5\%$). The SI axis behaves exactly as before---a column
effect, large for the weak Qwen verifier on both bases ($+8.7$, $+18.5$) and near zero for
the strong Muse verifier ($-1.1$, $+0.5$). Beating self-consistency, however, becomes a
row effect set by base saturation: on the Qwen-base pool every verifier matches or beats
the $79.2\%$ majority, and the strong verifier wins outright ($+7.94$), whereas on the
near-saturated Muse-base pool ($87.5\%$ majority) neither verifier does. Because the task
is held fixed and only the pool changes---the base model together with its frame budget---this contrast tracks saturation the way
Video-Holmes isolates strength.

\paragraph{A positive increment need not beat self-consistency.} Positive SI means the
candidates help the verifier beyond its own solving; it does not guarantee beating computationally cheap
self-consistency, above all when self-consistency is already near its ceiling. VRBench is the
clearest case. At the suite's lowest headroom both verifiers post positive SI (Muse $+3.67$,
GPT $+3.96$) yet both significantly trail self-consistency ($\Delta_{\text{maj}}{=}-3.26$,
$p{=}5.6\times10^{-6}$; $-1.62$, $p{=}3.7\times10^{-3}$), because self-consistency there is
near ceiling ($\approx91\%$), which leaves the verifier little to gain and easy ground to
lose. MLVU-Test with the GPT verifier repeats the pattern with a much larger increment
($\mathrm{SI}{=}+11.16$) and $\Delta_{\text{maj}}{=}-1.99$ (n.s.). Figure~\ref{fig:taxonomy} places all deployed benchmark--verifier pairs in these two coordinates: the stronger verifier raises SI
everywhere, but whether a pair clears the self-consistency line is a separate matter---some
rise into the useful quadrant, while VRBench and MLVU-Test stay below it despite large
positive increments.

\paragraph{A mechanistic reading, and a benchmark-specific inversion.} The lever behind SI is
a verifier's own standalone accuracy on the benchmark, measured independently of any selection
outcome: SI is large when that accuracy is low and collapses as it rises. Because this is not
raw capability---the same model can be a weak solver on one benchmark and a strong one on
another---it predicts a benchmark-specific inversion inside the InternVL3 family. On EgoSchema
the increment \emph{rises} from the $8$B to the $\sim\!5\times$-larger $38$B verifier ($+0.4$,
n.s., $\to +5.6$, $p{=}0.01$), because there the $38$B actually under-solves the $8$B
(standalone $72.4\%$ vs.\ $79.2\%$---the $8$B out-solves its larger sibling on this
benchmark), leaving the larger verifier more room; the standalone ordering reverts on
MLVU-Test ($8$B $51.1\%$ vs.\ $38$B $62.0\%$), where the increment falls with strength as
usual. Consistent with the inverse rule, the weakest solver posts the largest increment on
both benchmarks---Phi-3.5's SI reaches $+19.8$ on VRBench and $+25.2$ on EgoSchema---and the
decoupling again runs both ways: on EgoSchema InternVL3-8B beats self-consistency ($+3.0$)
despite a negligible increment ($\mathrm{SI}{=}+0.4$, n.s.). Two levers, then: a verifier's
independently measured standalone accuracy governs the \emph{size} of the Selection Increment
(inversely), while its strength---with base difficulty held fixed, as on Video-Holmes---governs
whether selection is worth more than a majority vote.

\subsection{Answer-position bias: mechanism, frontier behavior, and prior work}

These paragraphs extend Section~\ref{section:position}.

The mechanism is a graded positional slope, not a first-option collapse. On Video-Holmes (verifier $K{=}24$) the weak Phi-3.5 verifier's per-letter deviation from gold runs monotonically from
$+4.25$ (A) through $+3.81$, $+2.34$, $+1.52$ down to $-4.41$ (E) and $-7.51$ (F)---a
range of $11.76$ points at a perfect rank order---and the last-option deficit ($-7.51$)
exceeds the first-option excess ($+4.25$) in magnitude; EgoSchema (verifier $K{=}48$) shows the same asymmetry
($+8.0$ at A to $-10.0$ at E, range $18.0$). This is why primacy, not letter ``A'', is
the right headline: across benchmarks the weak verifier's primacy is positive on all
five (within-benchmark $\rho$ with accuracy $=-0.900$, Fisher $p{=}0.0027$), whereas its
bare A-lift is positive on only four---on TempCompass the verifier still drifts toward
earlier options yet its A-lift is flat ($0.0$), so a letter-``A'' statistic alone would
miss the bias that primacy detects. One caveat: because every benchmark
here labels options alphabetically, an option's position in the list and its letter rank
move together, so neither primacy nor A-lift, as we report them, separates a preference for
earlier positions from a preference for earlier letters; separating them cleanly would
require re-running the verifier with the option labels permuted against fixed
slots---merely reordering the options leaves ``A'' in the first slot and does not break the
collinearity---which we leave to future work.

At the frontier the pattern narrows but does not reverse, and even the narrowing is
partly a property of the candidate pool. Our two deployed verifiers (Muse and GPT-5.6
Sol) track gold on TempCompass, MLVU-Test, and VRBench (A-lift between $-0.13$ and
$-1.8$). Where a strong verifier's A-lift turns negative it is inheriting an A-poor
candidate pool---the sampled candidates under-represent ``A'' relative to gold by
$1.36$--$1.66$ points on Video-Holmes, MLVU-Test, and TempCompass. Measured against the
pool the verifier is actually offered, no frontier verifier is anti-``A'' (Video-Holmes
GPT-5.6 Sol $-1.20\!\to\!+0.26$, Muse $-0.88\!\to\!+0.54$), and the frontier verifiers'
own candidate-free A-lift is positive (GPT $+1.74$, Muse $+1.15$); the frontier is thus
statistically indistinguishable from zero, not negative. Recency avoidance, by contrast,
persists undiminished at every capability level: the last-option deficit is positive in
$21$ of $22$ cells (sole exception MLVU-Test/InternVL3-38B, $-0.80$), tracks accuracy at
within-benchmark $\rho{=}-0.900$ (Fisher $p{=}0.0025$), and for nine cells---all but one
on EgoSchema---the deficit exceeds the first-option excess (EgoSchema Muse $+4.60$). The
deficit persists even at the frontier, where the first-option excess is absent (Video-Holmes GPT $+1.58$, Muse $+0.60$). Position bias narrows at the frontier but does not disappear.

We report significance conservatively, and the statistics support a threshold rather than
a law. Every permutation $p$ in this section is one-sided against the predicted negative
direction; the ten-configuration test enumerates all $10!$ label assignments exactly, while
the pooled within-benchmark tests are Monte-Carlo estimates over $2\times10^{5}$ draws.
Pooled across all cells with benchmark fixed effects, the accuracy--skew rank
correlation is strong ($\rho{=}-0.8313$, within-benchmark permutation
$p{\approx}8\times10^{-5}$), but the attenuation just described shows the under-capable end
carrying it. Of the $22$ cells, nine are individually significant by an exact
McNemar test ($\alpha{=}0.05$, uncorrected) and five survive Holm--Bonferroni correction.
All four \emph{positive} survivors belong to Phi-3.5-Vision; the fifth is the capable
Qwen3.5-27B self-verifier on Video-Holmes, which clears correction with its sign
\emph{reversed} ($p{=}0.0020$, $-2.8$ points)---below gold, not above it. The bias that
survives correction is thus entirely a weak-verifier effect, and the capable verifier
supplies a signed control on it. Two qualifications keep the claim narrow. The mid-tier InternVL3 family is
genuinely anti-``A'' even against the offered pool---more so than the frontier closed
models---so we claim capability gating and not an ordering of capability tiers by raw bias
means. And VRBench, run on two verifiers, contributes only a sign, its $k{=}2$ rank
correlation having a floor exact $p$ of $0.5$. The weak-verifier default is also
benchmark-dependent (Table~\ref{table:alift}): Phi-3.5 lifts ``A'' by $+8.6$ on MLVU-Test
and $+3.0$ on VRBench, while on TempCompass no verifier defaults at all. The capable
self-verifier does not default on either cell we could re-measure without a truncating
generation cap (MLVU-Test $-2.2$, Video-Holmes $-2.8$); its apparent default in the
shipped measurement was an artifact of that cap.

Answer-position bias is a well-documented failure mode of language-model selectors: an
inherent behavioral bias over option-ID tokens rather than a prompting
artifact~\citep{zheng2024mcselectors,robinson2023mcsb}, it swings accuracy under option
reordering~\citep{pezeshkpour2023options}, destabilizes LLM-as-a-judge
evaluation~\citep{zheng2023judging,wang2024fairevaluators,shi2024judgingjudges}, and has
been measured in video- and vision-language multiple-choice
models~\citep{loginova2024blindguessing,tan2024ordermatters,atabuzzaman2025lvlmselection}.
That capability attenuates it is likewise known: stronger vision-language models show
reduced but nonzero selection bias~\citep{atabuzzaman2025lvlmselection}, accuracy gates
rather than removes length-driven position bias in reasoning
models~\citep{wang2026morethinking}, and more capable judges hold up better as candidate
lists grow~\citep{shi2024judgingjudges}. On that axis our finding is a replication in a new
setting. What has not been measured is position bias in the \emph{best-of-$N$ verifier}
role---selecting among a model's own sampled candidates rather than answering a
multiple-choice item or ranking a pair. There the skew belongs to the verifier and not to
the surrounding pipeline: the candidate generator is B/C-skewed rather than A-skewed, and
the majority-vote tie-break is exactly insertion-order rather than alphabetical, which
rules out the argmax-tie mechanism identified for prior video-language
selectors~\citep{loginova2024blindguessing}. Our contribution is that setting and its
controls, not a new phenomenon---frontier judges can be reliable yet
position-biased~\citep{norman2026reliability}, bias direction is
model-specific~\citep{xu2026pointwisepairwise}, and near-ceiling accuracy can suppress
measurable bias for reasons unrelated to robustness~\citep{tamba2026ceiling}.

\section{When a verifier pass is worth its compute}
\label{appendix:cost}

Section~\ref{section:bounds} compresses our computational cost reasoning into a three-step allocation
policy. Because that policy is the part of this study a practitioner is most likely to act
on, we set it out at length here, and separate what we measured from what we infer from the
measurements.

\subsection{What the second pass requires, and what it does not}

The accounting is easiest to state in \emph{passes} rather than tokens. Sampling $N$
candidates from the base is a requirement of admission for every method considered here,
including the baseline: once those candidates exist, self-consistency is computationally cheap, because a
majority vote reads the answer letters and never calls a model again. Verification adds one
listwise pass in which the verifier re-watches the footage and reads all $N$ candidates, so
it is the only step whose computational cost scales with both the video and the candidate set. The
diagnostic we recommend adds one further pass---the verifier answering the same question
with the candidates withheld---which is bounded by the computational cost of the verifier's own solving
and, unlike the selection pass, does not grow with $N$.

We deliberately do not convert this into a token or wall-clock figure. Our runs were not
instrumented for a benchmark-matched token accounting, both strong verifiers are reached by
API under serving conditions we do not control, and frame budgets differ across rungs
(Appendix~\ref{appendix:extended}). Any efficiency ratio we printed would therefore report
provisioning rather than a property of the method, so we make no compute-efficiency claim
and treat computational cost qualitatively throughout.

\subsection{Why the computational overhead can be justified}

\paragraph{Diagnosis is computationally cheaper than the treatment.} The two gates do not require the same amount of compute, and the computationally cheaper one is decisive wherever it applies. Majority accuracy over the
sampled candidates requires nothing beyond the candidates themselves, so a practitioner can
discover that verification is the wrong decision \emph{before} choosing any of it: above
roughly $87\%$ no verifier in our study beat the vote, and on VRBench both deployed
verifiers significantly lost to it ($-1.62$ and $-3.26$). Only once that computationally cheap gate is
passed does the computationally expensive diagnostic become relevant---and it is the computationally expensive one, a single pass
with the candidates withheld, that tells apart the two systems a headline gain conflates.

\paragraph{The increment separates two computational costs that one accuracy number merges.} Read as
an audit rather than a merit score, the sign of the increment answers a question about
computational cost rather than about quality. Where it is negative---TempCompass with the Muse
verifier, $\mathrm{SI}{=}-1.08$---the verifier is more accurate answering directly than
selecting, so the pipeline incurs computational cost for the candidate set and returns less than the verifier
alone would. Where it is positive---MLVU-Test with that same verifier, $+5.68$---selection
provides something standalone solving does not supply, and it does so even in cases where the
result still fails to clear a computationally cheap majority vote. Neither fact is recoverable from the gain
over the vote, which is why we report the two axes together.

\paragraph{A weak verifier can be a computationally expensive way to reproduce a computationally cheap baseline.} The
sharpest form of the computational cost argument never appears in an accuracy column. On the Qwen-base
TempCompass pool the weak self-verifier agrees with the plurality candidate on $95.1\%$ of
rows, so its selection is close to a computationally expensive re-derivation of the vote it was meant to
improve on: the pass is spent, the footage is re-watched, and the modal answer is returned.
A delta near zero reads as harmless, but in computational cost terms it is the least attractive outcome
available, because of the compute required for output the majority vote supplied for nothing.

\paragraph{Reliability, not only accuracy.} Below a capability threshold the computational overhead also
provides a failure mode. The weakest verifiers track where an option sits in the
list---favoring earlier positions and avoiding the last---while every capable verifier we
measured resists this (Section~\ref{section:position}). That matters for deployment in a
way an averaged accuracy hides: a positional preference is a systematic error correlated
with option order, so it does not average away over a benchmark, and it will track whatever
ordering convention a downstream system happens to adopt. A verifier that merely matches
the vote, but reaches that accuracy partly through option order, is a weaker component than
the arithmetic suggests.

\paragraph{Sampling widens the gap the vote cannot reach.} One argument does favor the computational overhead of selection, and it concerns how the two quantities move with $N$. On Video-Holmes,
raising $N$ from $8$ to $16$ moves the best-of-$N$ oracle from $78.8\%$ to $82.5\%$ while
self-consistency stays flat ($58.9\%$ to $59.0\%$; Table~\ref{table:vhholmes}). Additional
samples enlarge what a selector could in principle win without improving the computationally cheap baseline
it has to beat, so the ceiling on selection is not pinned by the base model in the way the
vote's is. We report this as a property of our anchor rather than a scaling law: two values
of $N$ establish a direction, not a rate, and the strongest verifier we ran still left
roughly $17$ points of that oracle headroom unclaimed.

\subsection{Where the computational overhead is not justified}

Three results bound the case for spending, and we state them as bounds rather than as
discouragements. First, near-saturated pools: where the vote already sits at the observed
ceiling, verification did not help and sometimes significantly hurt, so the allocation is
to skip it. Second, a non-positive increment: there the verifier should be deployed as the
solver directly, which is both more accurate and computationally cheaper, since it avoids reading the
candidates at all. Third, the obvious alternative expenditure does not rescue the
situation---augmenting candidate generation with an on-demand loop that may zoom, crop,
retrieve frames, or ground objects left the best-of-$N$ oracle unchanged
($\Delta_{\text{oracle}}{=}0.000$; Appendix~\ref{appendix:control}), so a practitioner
facing a pool that selection cannot exploit cannot repair it by gathering more evidence
into that same pool. What remains, on our anchor, is the narrow case the body identifies: a
pool below the saturation ceiling, paired with a verifier clearly stronger than the base
and drawn from a different model family.

\end{document}